\documentclass[10pt, conference, compsocconf]{IEEEtran}
\usepackage{cite}

\usepackage[numbers]{natbib}

\ifCLASSINFOpdf
  \usepackage[pdftex]{graphicx}
  \graphicspath{{../pdf/}{../jpeg/}}
  \DeclareGraphicsExtensions{.pdf,.jpeg,.png}
\else
  \usepackage[dvips]{graphicx}
  \graphicspath{{../eps/}}
  \DeclareGraphicsExtensions{.eps}
\fi
\usepackage[cmex10]{amsmath}
\usepackage{algorithm}
\usepackage[algo2e]{algorithm2e}
\usepackage[caption=false,font=footnotesize]{subfig}
\usepackage{subfloat}
\usepackage{enumerate}
\usepackage{amsmath}
\usepackage{amssymb}
\usepackage{multirow}
\usepackage[flushleft]{threeparttable}

\begin{document}

\title{Embedding Feature Selection for Large-scale Hierarchical Classification}

\author{\IEEEauthorblockN{Azad Naik and Huzefa Rangwala}
\IEEEauthorblockA{
Department of Computer Science\\
George Mason University\\
Fairfax, VA, United States\\
Email: anaik3@gmu.edu, rangwala@cs.gmu.edu}
}
\maketitle

\begin{abstract}
Large-scale Hierarchical Classification (HC) involves datasets consisting of thousands of classes and millions of training instances with high-dimensional features posing several big data challenges. Feature selection that aims to select the subset of discriminant features is an effective strategy to deal with large-scale HC problem. It speeds up the training process, reduces the prediction time and minimizes the memory requirements by compressing the total size of learned model weight vectors. Majority of the studies have also shown feature selection to be competent and successful in improving the classification accuracy by removing irrelevant features. In this work, we investigate various filter-based feature selection methods for dimensionality reduction to solve the large-scale HC problem. Our experimental evaluation on text and image datasets with varying distribution of features, classes and instances shows upto 3x order of speed-up on massive datasets and upto 45$\%$ less memory requirements for storing the weight vectors of learned model without any significant loss (improvement for some datasets) in the classification accuracy. Source Code: https://cs.gmu.edu/$\sim$mlbio/featureselection.
\end{abstract}

\begin{IEEEkeywords}
Feature Selection, Top-down Hierarchical Classification, Logistic Regression, Scalability
\end{IEEEkeywords}

%
\IEEEpeerreviewmaketitle

\section{Introduction}
\label{intro}
Hierarchies (Taxonomies) are popular for organizing large volume datasets in various application domains \cite{ashburner2000gene,deng2009imagenet}. Several large-scale online prediction challenges such as LSHTC\footnote{http://lshtc.iit.demokritos.gr/} (webpage classification), BioASQ\footnote{http://bioasq.org/} (PubMed documents classification) and ILSVRC\footnote{http://www.image-net.org/challenges/LSVRC/2016/} (image classification) revolve around the HC problem. Although, substantial amount of data with inter-class dependencies information are beneficial for improving HC, one of the major challenges in dealing with these datasets comprising large-number of categories (classes), high-dimensional features and large-number of training instances (examples) is {scalability}.

Many large-scale HC approaches have been developed in past to deal with the various ``big data" challenges by: (i) training faster models, (ii) quickly predicting class-labels and (iii) minimizing memory usage. For example, Gopal et al. \cite{gopal2013distributed} proposed the log-concavity bound that allows parallel training of model weight vectors across multiple computing units. This achieves significant speed-up along with added flexibility of storing model weight vectors at different units. However, the memory requirements is still large ($\sim$26 GB for DMOZ-2010 dataset, refer to Table \ref{table:paramsize}) which requires complex distributed hardware for storage and implementation. Alternatively, Map-Reduce based formulation of learning model is introduced \cite{naik2015ranking,gopal2013recursive} which is scalable but have software/hardware dependencies that limits the applicability of this approach. 

To minimize the memory requirements, one of the popular strategy is to incorporate the feature selection in conjunction with model training \cite{zhou2011hierarchical,heisele2003hierarchical}. The main intuition behind these approaches is to squeeze the high-dimensional features into lower dimensions. This allows the model to be trained on low-dimensional features only; significantly reducing the memory usage while retaining (or improving) the classification accuracy. This is possible because only subset of features are beneficial to discriminate between classes at each node in the hierarchy. For example, to distinguish between sub-class `Chemistry' and `Physics' that belongs to class `Science' features like \emph{chemical}, \emph{reactions} and \emph{acceleration} are important whereas features like \emph{coach}, \emph{memory} and \emph{processor} are irrelevant. HC methods that leverage the structural relationship shows improved classification performance but are computationally expensive \cite{liu2005support,gopal2013recursive,cai2004hierarchical}. 

\begin{figure*}
\centering
  \includegraphics[width=1\linewidth,height=7.5cm]{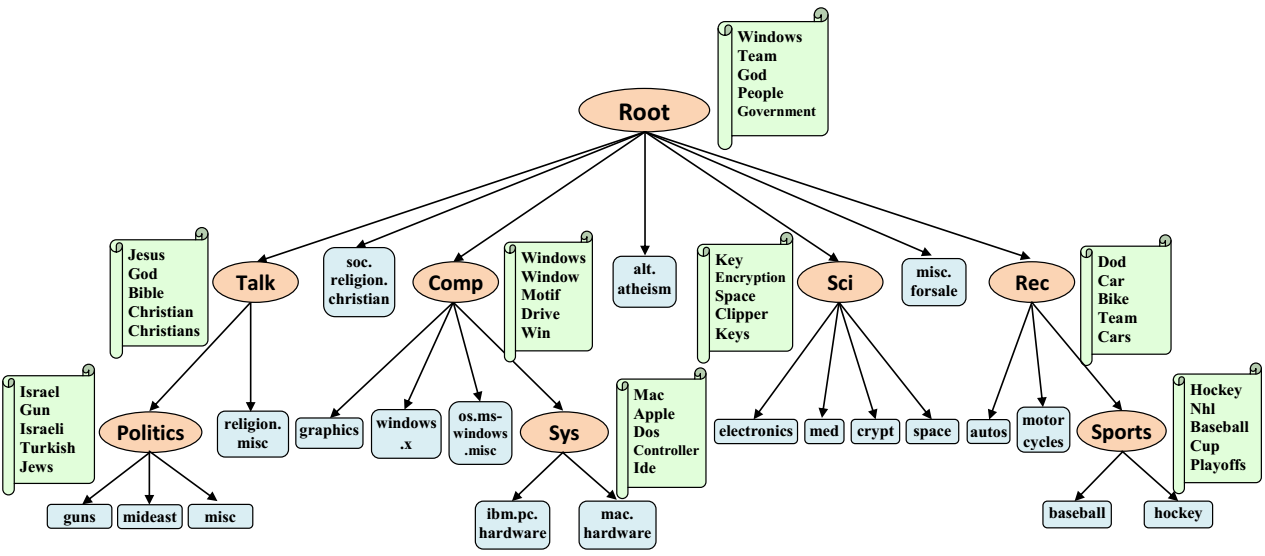}
  \caption{\textbf{Figure demonstrating the importance of feature selection for HC. Green color (sticky note) represents the top five best features selected using gini-index feature selection method at each internal node. Internal nodes are represented by orange color (elliptical shape) and leaf nodes are represented by blue color (rectangular shape).}}
\label{FSfigure}
\end{figure*}

In this paper, we study different filter-based feature selection methods for solving large-scale HC problem. Feature selection serves as the preprocessing step in our learning framework prior to training models. Any developed methods for solving HC problem can be integrated with the selected features, providing flexibility in choosing the HC algorithm of our choice along with computational efficiency and storage benefits. Our proposed ``adaptive feature selection" also shows an improvement of $\sim$2\% in classification accuracy. Experiments on various real world datasets across different domains demonstrates the utility of the feature selection over full set of high-dimensional features. We also investigate the effect of feature selection in classification performance when the number of labeled instances per class is low.

\section{Literature Review}
\label{sec:1}
\subsection{Hierarchical Classification}
Several methods have been developed to address the hierarchical classification problem \cite{naik2015ranking,gopal2013recursive,naik2016inconsistent,dumais2000hierarchical,cesa2006incremental}. These methods can be broadly divided into three major categories. (i) \emph{Flat approach} is one of the simplest and straight forward method to solve the HC problem. In this method, hierarchical structure is completely ignored and an independent one-vs-rest or multi-class classifiers are trained for each of the leaf categories that can discriminate it from remaining leaf categories. For predicting the label of instances, the flat method invokes the classifiers corresponding to all leaf categories and selects the leaf category with highest prediction score. As such, flat approach have expensive training and prediction runtime for datasets with large number of classes. (ii) \emph{Local classification} involves the use of local hierarchical relationships during the model training process. Depending on how the hierarchical relationships are leveraged, various local methods exist \cite{silla2011survey}. In this paper, we have used the most popular ``local classifier per parent node" method as the baseline for evaluations. Specifically, we train a multi-class classifier at each of the parent node to maximize the discrimination between its children nodes. For making predictions a top-down method (discussed in Section \ref{topdown}) is followed. (iii) \emph{Global classification} learns a single complex model where all relevant hierarchical relationships are explored jointly during the optimization process making these approaches expensive for training. Label predictions is done using a similar approach followed for flat or local methods.

\subsection{Top-Down Hierarchical Classification}
\label{topdown}
One of the most efficient approach for solving large-scale HC problem is using the top-down method \cite{liu2005support,koller1997hierarchically}.
In this method, local or global classification method is used for model training and the unlabeled instances are recursively classified in a top-down fashion. At each step, best node is picked based on the computed prediction score of its children nodes. The process repeats until the leaf node representing a certain category (or class-label) is reached, which is the final predicted label (refer to eq. (\ref{eq:TDTest})). 

Top-Down (TD) methods are popular for large-scale problems owing to their computational benefits where only the subset of classes in the relevant path are considered during prediction phase. For example, in order to make second level prediction provided the first level prediction is `{Sci}' (shown in Figure \ref{FSfigure}) we only need to consider the children of `{Sci}' class ($i.e.$, {electronics}, {med}, {crypt} and {space}), thereby, avoiding the large number of second level classes such as `{Sys}', `Politics', `Sports', `graphics', `autos'. In the past, top-down methods have been  successfully used to solve HC problems \cite{dumais2000hierarchical,holden2005hybrid,li2005music}. Liu et al. \cite{liu2005support} performed classification on large-scale Yahoo! dataset and analyzed the complexity of the top-down approach. In \cite{secker2007experimental}, a selective classifier top-down method is proposed where the classifier to train at particular node is chosen in a data-driven manner.

\subsection{Feature Selection}
There have been several studies focused on feature selection methods for the flat classification problem \cite{tang2014feature,dash1997feature,peng2005feature,zheng2004feature,joachims1998text,kohavi1997wrappers}. However, very few work emphasize on feature selection for HC problem that are limited to small number of categories \cite{ristoski2014feature,wibowo2002simple}. Figure \ref{FSfigure} demonstrates the importance of feature selection for hierarchical settings where only the relevant features are chosen at each of the decision (internal) nodes. More details about the figure will be discussed in Section \ref{resDiscuss} (Case Study).

Feature selection aims to find a subset of highly discriminant features that minimizes the error rate and improve the classifier performance. Based on the approach adapted for selecting features two broad categories of feature selection exist, namely, wrapper and filter-based methods. Wrapper approaches evaluate the fitness of the selected features using the intended classifier. Although many different wrapper-based approaches have been proposed, these methods are not suitable for large-scale problems due to the expensive evaluation needed to select the subset of features \cite{tang2014feature}. On the contrary, filter approaches select the subset of features based on the certain measures or statistical properties that does not require the expensive evaluations. This makes the filter-based approaches a natural choice for large-scale problem. Hence, in this paper we have focused on various filter-based approaches for solving HC problem (discussed in Section \ref{filterFeatSelect}). In literature, third category referred as embedded approaches have also been proposed which are a hybrid of the wrapper and filter methods. However, these approaches have not been shown to be efficient for large-scale classification \cite{tang2014feature} and hence, we do not focus on hybrid methods.

To the best of our knowledge this is the first paper that performs a broad study of filter-based feature selection methods for HC problem.

\section{Methods}
\label{featMethod}
\subsection{Definitions and Notations}
In this paper, we use {\bf{bold}} lower-case and upper-case letters to indicate vector and matrix variables, respectively. Symbol $\mathcal{N}$ denotes the set of internal nodes in the hierarchy where for each node $n \in \mathcal{N}$ we learn the multi-class classifier denoted by ${\bf{W}}_n$ = $\big[{\bf{w}}^n_c\big]_{{c \in \mathcal{C}(n)}}$ to discriminate between its children nodes $\mathcal{C}(n)$. ${\bf{w}}^n_c$ represents the optimal model weight vectors for $c^{th}$ child of node $n$. $\mathcal{L}$ denotes the set of leaf nodes (categories) to which instances are assigned. The total number of training instances are denoted by $N$ and $T(n) \subseteq N$ denotes the total number of training instances considered at node $n$ which corresponds to all instances of descendant categories at node $n$. $\mathcal{F}$ denotes the set of total features (dimensionality) for each instance where $i^{th}$ feature is denoted by $f_i$. ${S}_\mathcal{F} \subseteq \mathcal{F}$ denotes the subset of relevant features selected using feature selection algorithm. $\mathcal{D} = \big\{\big({\bf{x}}(i)$, $y(i)\big)\big\}_{i=1}^{N}$ denotes the training dataset where ${\bf{x}}(i) \in \mathbb{R}^{\mathcal{F}}$ and $y(i) \in \mathcal{L}$. For training optimal model corresponding to $c^{th}$ child at node $n$ we use the binary label ${y_c^n}{(i)} \in \{\pm1\}$ for $i^{th}$ training instance where $y_c^n(i)$ = 1 iff $y(i)$ = $c$ and $y_c^n(i)$ = -1 otherwise. Predicted label for $i^{th}$ test instance $\hat{\bf{x}}(i)$ is denoted by $\hat{y}(i) \in \mathcal{L}$. 

\subsection{Hierarchical Classification}
\label{HierClass}
Given a hierarchy $\mathcal{H}$, we train multi-class classifiers for each of the internal nodes $n\in\mathcal{N}$ in the hierarchy--- to discriminate between its children nodes $\mathcal{C}(n)$. In this paper, we have used Logistic Regression (LR) as the underlying base model for training \cite{gopal2013recursive,naik2013classifying}. The LR objective uses logistic loss to minimize the empirical risk and $l_1$-norm (denoted by $\big|\big|\cdot\big|\big|_{1}$) or squared $l_2$-norm term (denoted by $\big|\big|\cdot\big|\big|_{2}^{2}$) to control model complexity and prevent overfitting. Usually, $l_1$-norm encourages sparse solution by randomly choosing single parameter amongst highly correlated parameters whereas $l_2$-norm jointly shrinks the correlated parameters. The objective function $\Psi^{n}_{c}$ for training a model corresponding to ${c^{th}}$ child of node $n$ is provided in eq. (\ref{ARLR}).
\begin{equation}
\scriptsize
\Psi^{n}_{c}=\min_{{\bf{w}}^{n}_{c}}\Bigg[\lambda\sum_{i=1}^{T(n)}\log\left(1+\exp\left(-y_{c}^{n}(i)\big({\bf{w}}^{n}_c\big)^{T}\mathbf{x}(i)\right)\right)+ \mathcal{R}\big({\bf{w}}^{n}_c\big)\Bigg]\label{ARLR}
\end{equation}

where $\lambda > 0$ is a mis-classification penalty parameter and $\mathcal{R}\big(.\big)$ denotes the regularization term given by eq. (\ref{regTerm}).
\begin{align}
\mathcal{R}\big({\bf{w}}^n_c\big)=\left\{ \begin{alignedat}{1} & \big|\big|{\bf{w}}^n_c\big|\big|_1^1, \quad l_1-norm\\
 & \quad \quad \quad OR\\
 & \big|\big|{\bf{w}}^n_c\big|\big|_2^2, \quad l_2-norm
\end{alignedat}
\right\} \label{regTerm}
\end{align}

For each child $c$ of node $n$ within the hierarchy, we solve eq. (\ref{ARLR}) to obtain the optimal weight vector denoted by ${\bf{w}}^{n}_c$. The complete set of parameters for all the children nodes ${\bf{W}}_n$ = $[{\bf{w}}^{n}_c] _{c\in\mathcal{C}(n)}$ constitutes the learned multi-class classifiers at node $n$ whereas total parameters for all internal nodes ${\bf{W}}$ = $[{\bf{W}}_n]_{n \in \mathcal{N}}$ constitutes the learned model for Top-Down (TD) classifier. 

For a test instance ${\hat{\bf{x}}}(i)$, the TD classifier predicts the class label $\hat{y}(i)\in\mathcal{L}$ as shown in eq. (\ref{eq:TDTest}). Essentially, the algorithm starts at the root node and recursively selects the best child nodes until it reaches a terminal node belonging to the set of leaf nodes $\mathcal{L}$.
\begin{align}
\hat{y}(i)=\left\{ \begin{alignedat}{1} & \mathbf{initialize}\quad p:=root\\
 & \mathbf{while}\ p\notin\mathcal{L}\\
 & \quad p:=\mathbf{argmax}_{q\in\mathcal{C}(p)}\ \big({\bf{w}}^p_{q}\big)^T\hat{{\bf{x}}}(i)\\
 & \mathbf{return}\ p
\end{alignedat}
\right\} \label{eq:TDTest}
\end{align}

\subsection{Feature Selection}
\label{filterFeatSelect}
The focus of our study in this paper is on filter-based feature selection methods which are scalable for large-scale datasets. In this section, we present four feature selection approaches that are used for evaluation purposes.\\
\\
{\textbf{Gini-Index}} - It is one of the most widely used method to compute the best split (ordered feature) in the decision tree induction algorithm \cite{ogura2009feature}. Realizing its importance, it was extended for the multi-class classification problem \cite{shang2007novel}. In our case, it measure the feature's ability to distinguish between different leaf categories (classes). Gini-Index of $i^{th}$ feature $f_i$ with $\mathcal{L}$ classes can be computed as shown in eq. (\ref{gi}).
\begin{equation}
{\bf{Gini-Index}}(f_i) = 1 - \sum_{k=1}^{\mathcal{L}}\Big(p(k|f_i)\Big)^2
\label{gi}
\end{equation}
where $p(k|f_i)$ is the conditional probability of class $k$ given feature $f_i$.
 
Smaller the value of Gini-Index, more relevant and useful is the feature for classification. For HC problem, we compute the Gini-Index corresponding to all feature's independently at each internal node and select the best subset of features (${S}_\mathcal{F}$) using a held-out validation dataset. 
\\
\\
{\textbf{Minimal Redundancy Maximal Relevance (MRMR)}} - This method incorporates the following two conditions for feature subset selection that are beneficial for classification. 
\begin{enumerate}[(i)]
\item Identify features that are mutually maximally dissimilar to capture better representation of entire dataset and
\item Select features to maximize the discrimination between different classes. 
\end{enumerate}
The first criterion referred as ``minimal redundancy" selects features that carry distinct information by eliminating the redundant features. The main intuition behind this criterion is that selecting two similar features contains no new information that can assist in better classification. Redundancy information of feature set $\mathcal{F}$ can be computed using eq. (\ref{minred}).
\begin{equation}
{\mathfrak{R}_D} =  \Bigg[ \frac{1}{|S_\mathcal{F}|^2}\sum_{f_i, f_j \in S_\mathcal{F}}I(f_i, f_j)\Bigg]
\label{minred}
\end{equation}
where $I(f_i, f_j)$ is the mutual information that measure the level of similarity between features $f_i$ and $f_j$ \cite{ding2005minimum}.

The second criterion referred as ``maximum relevance'' enforces the selected features to have maximum discriminatory power for classification between different classes. Relevance of feature set $\mathcal{F}$ can be formulated using eq. (\ref{maxrel}).
\begin{equation}
{\mathfrak{R}_L} = \Bigg[ \frac{1}{|S_\mathcal{F}|}\sum_{f_i\in S_\mathcal{F}}I(f_i, \mathcal{L})\Bigg]
\label{maxrel}
\end{equation} 
where $I(f_i, \mathcal{L})$ is the mutual information between the feature $f_i$ and leaf categories $\mathcal{L}$ that captures how well the feature $f_i$ can discriminate between different classes \cite{peng2005feature}.

The combined optimization of eq. (\ref{minred}) and eq. (\ref{maxrel}) leads to a feature set with maximum discriminatory power and minimum correlations among features. Depending on strategy adapted for optimization of these two objectives different flavors exist. The first one referred as ``mutual information difference (MRMR-D)" formulates the optimization problem as the difference between two objectives as shown in eq. (\ref{mrmrd}). The second one referred as ``mutual information quotient (MRMR-Q)" formulates the problem as the ratio between two objectives and can be computed using eq. (\ref{mrmrq}).
\begin{equation}
\emph{MRMR-D} = \max_{S_\mathcal{F} \subseteq \mathcal{F}} \; ({{\mathfrak{R}_L - \mathfrak{R}_D}})
\label{mrmrd}
\end{equation} 
\begin{equation}
\emph{MRMR-Q} = \max_{S_\mathcal{F} \subseteq \mathcal{F}} \; ({{\mathfrak{R}_L/\mathfrak{R}_D}})
\label{mrmrq}
\end{equation} 
For HC problem again we select the best top ${S}_\mathcal{F}$ features (using a validation dataset) for evaluating these methods.
\\
\\
{\textbf{Kruskal-Wallis}} - This is a non-parametric statistical test that ranks the importance of each feature. As a first step this method ranks all instances across all leaf categories $\mathcal{L}$ and computes the feature importance metric as shown in eq. (\ref{kw}):
\begin{equation}
KW=(N-1)\frac{\sum_{i=1}^{\mathcal{L}}{n_i(\bar{r_i}-\bar{r})^2}}{\sum_{i=1}^{\mathcal{L}}\sum_{j=1}^{n_i}n_i({r}_{ij}-\bar{r})^2}
\label{kw}
\end{equation}
where $n_i$ is the number of instances in $i^{th}$ category, $r_{ij}$ is the ranking of $j^{th}$ instances in the $i^{th}$ category and $\bar{r}$ denotes the average rank across all instances.

It should be noted that using different feature results in different ranking and hence feature importance. Lower the value of computed score $KW$, more relevant is the feature for classification.

\begin{algorithm}[t!]
\SetAlgoLined
 \KwData{Hierarchy $\mathcal{H}$, input-output pairs \Big(${\bf{x}}(i), y(i)$\Big)}
 \KwResult{Learned model weight vectors:

\hspace*{11mm} ${\bf{W}}$ = [${\bf{W}}_1$, ${\bf{W}}_2$, $\cdots$, ${\bf{W}}_n$], $n$ $\in$ $\mathcal{N}$}

${\bf{W}}$ = $\phi$;

/* \textbf{1st subroutine: Feature Selection} */

  \For{$f_i \in$ $\mathcal{F}$}{
      Compute score (relevance) corresponding to feature $f_i$ using feature selection algorithm mentioned in Section \ref{filterFeatSelect};
   }
   
   Select top $k$ features based on score (and correlations) amongst features where best value of $k$ is tuned using a validation dataset

/* \textbf{2nd subroutine: Model Learning using Reduced Feature Set} */

  \For{n $\in$ $\mathcal{N}$}{
     /* \textbf{learn models for discriminating child at node $n$} */

      Train optimal multi-class classifiers ${\bf{W}}_n$ at node $n$ using reduced feature set as shown in eq. (\ref{ARLR});

      /* \textbf{update model weight vectors} */

      ${\bf{W}}$ = [${\bf{W}}$, ${\bf{W}}_n$];
   }

\Return{${\bf{W}}$}
 \caption{Feature Selection (FS) based Model Learning for Hierarchical Classification (HC)}
 \label{featSelectHC}
\end{algorithm}

\begin{table*}[tb] 
\centering 
\caption{\textbf{Dataset Statistics}} \label{table:finaltabledataset} 
\begin{tabular}{|l|  c c c  c c c c c|} 
\hline
\multirow{2}{*}{\bf{Dataset}}& \multirow{2}{*}{\bf{Domain}} & \multirow{2}{*}{\bf{\# Leaf Node}} & \multirow{2}{*}{\bf{\# Internal Node}} & \multirow{2}{*}{\bf{Height}} & \multirow{2}{*}{\bf{\# Training}} & \multirow{2}{*}{\bf{\# Testing}} & \multirow{2}{*}{\bf{\# Features}} & {\bf{Avg. \# children}}\\
\multicolumn{1}{|c|}{} & \multicolumn{7}{c}{} & (per internal node)\\
\hline
{\bf{NG}} & Text & 20 & 8 & 4 & 11,269 & 7,505 & 61,188 & 3.38\\
{\bf{CLEF}} & Image & 63 & 34 & 4 & 10,000 & 1,006 & 80 & 2.56\\
{\bf{IPC}} & Text & 451 & 102 & 4 & 46,324 & 28,926 & 1,123,497 & 5.41\\
{\bf{DMOZ-SMALL}} & Text & 1,139 & 1,249 & 6 & 6,323 & 1,858 & 51,033 & 1.91\\
{\bf{DMOZ-2010}} & Text & 12,294 & 4,928 & 6 & 128,710 & 34,880 & 381,580 & 3.49\\
{\bf{DMOZ-2012}} & Text & 11,947 & 2,016 & 6 & 383,408 & 103,435 & 348,548 & 6.93\\
\hline
\end{tabular}
\end{table*} 

\subsection{Proposed Framework}
Algorithm \ref{featSelectHC} presents our proposed method for embedding feature selection into the HC framework. It consist of two independent main subroutines: (i) a feature selection algorithm (discussed in Section \ref{filterFeatSelect}) for deciding the appropriate set of features at each decision (internal) node and (ii) a supervised learning algorithm (discussed in Section \ref{HierClass}) for constructing a TD hierarchical classifier using reduced feature set. Feature selection serves as the preprocessing step in our framework which provides flexibility in choosing any HC algorithm. 

We propose two different approaches for choosing relevant number of features at each internal node $n \in \mathcal{N}$. The first approach which we refer as ``global feature selection (Global FS)"  selects the same number of features for all internal nodes in the hierarchy where the number of features are determined based on the entire validation dataset performance. The second approach, referred as ``adaptive feature selection (Adaptive FS)" selects different number of features at each internal node to maximize the performance at that node. It should be noted that adaptive method only uses the validation dataset that exclusively belongs to the internal node $n$ ($i.e.$, descendant categories of node $n$). Computationally, both approaches are almost identical because model tuning and optimization requires similar runtime which accounts for the major fraction of computation. 

\section{Experimental Evaluations}
\subsection{Dataset Description}
We have performed an extensive evaluation of various feature selection methods on a wide range of hierarchical text and image datasets. Key characteristics about the datasets that we have used in our experiments are shown in Table \ref{table:finaltabledataset}. All these datasets are single-labeled and the instances are assigned to the leaf nodes in the hierarchy. For text datasets, we have used the word-frequency representation and perform the tf-idf transformation with $l_2$-norm to the word-frequency feature vector.\\
\textbf{Text Datasets}\\
 \hspace*{4mm}\textbf{NEWSGROUP (NG)}\footnote{http://qwone.com/$\sim$jason/20Newsgroups/} - It is a collection of approximately 20,000 news documents partitioned (nearly) evenly across twenty different topics such as `baseball', `{electronics}' and `{graphics}' (refer to Figure \ref{FSfigure}).\\
 \hspace*{4mm}\textbf{IPC}\footnote{http://www.wipo.int/classifications/ipc/en/} - Collection of patent documents organized in International Patent Classification (IPC) hierarchy.\\
 \hspace*{4mm}\textbf{DMOZ-SMALL, DMOZ-2010 and DMOZ-2012}\footnote{http://dmoz.org} - Collection of multiple web documents organized in various classes using the hierarchical structure. Dataset has been released as the part of the LSHTC\footnote{http://lshtc.iit.demokritos.gr/} challenge in the year 2010 and 2012. For evaluating the DMOZ-2010 and DMOZ-2012 datasets we have used the provided test split and the results reported for this benchmark is blind prediction obtained from web-portal interface\footnote{http://lshtc.iit.demokritos.gr/node/81}.\\
\textbf{Image Datasets}\\
 \hspace*{4mm}\textbf{CLEF} \cite{dimitrovski2011hierarchical} - Dataset contains medical images annotated with Information Retrieval in Medical Applications (IRMA) codes. Each image is represented by the 80 features that are extracted using local distribution of edges method.

\subsection{Evaluation Metrics}
We have used the standard set based performance measures Micro-$F_1$ and Macro-$F_1$ \cite{yang1999evaluation} for evaluating the performance of learned models.\\ 
\hspace*{4mm}\textbf{Micro-$F_1$ ($\mu$$F_1$)} - To compute $\mu$$F_1$, we sum up the category specific true positives $(TP_l)$, false positives $(FP_l)$ and false negatives $(FN_l)$ for different leaf categories and compute the $\mu$$F_1$ score as follows:
\begin{gather}P = \frac{\sum_{l \in \mathcal{L}}TP_l}{\sum_{l \in \mathcal{L}}(TP_l + FP_l)},
 R = \frac{\sum_{l \in \mathcal{L}}TP_l}{\sum_{l \in\mathcal{L}}(TP_l + FN_l)} \nonumber\\
\mu F_1 = \frac{2PR}{P + R}\end{gather}
where, P and R are the overall precision and recall values for all the classes taken together.\\
\hspace*{4mm}\textbf{Macro-$F_1$ ($M$$F_1$)} - Unlike $\mu$$F_1$, M$F_1$ gives equal weight to all the categories so that the average score is not skewed in favor of the larger categories. It is defined as follows: 
\begin{gather}P_l = \frac{TP_l}{TP_l + FP_l},
R_l = \frac{TP_l}{TP_l + FN_l} \nonumber\\ 
MF_1 = \frac{1}{|\mathcal{L}|}\sum_{l \in \mathcal{L}}\frac{2P_lR_l}{P_l + R_l}\end{gather}\\
where $|\mathcal{L}|$ denotes the set of leaf categories, $P_l$ and $R_l$ are the precision and recall values for leaf category $l$ $\in$ $\mathcal{L}$.

\begin{figure*}
	\center
        \subfloat[]{
            \includegraphics[width=.225\linewidth,height=4cm]{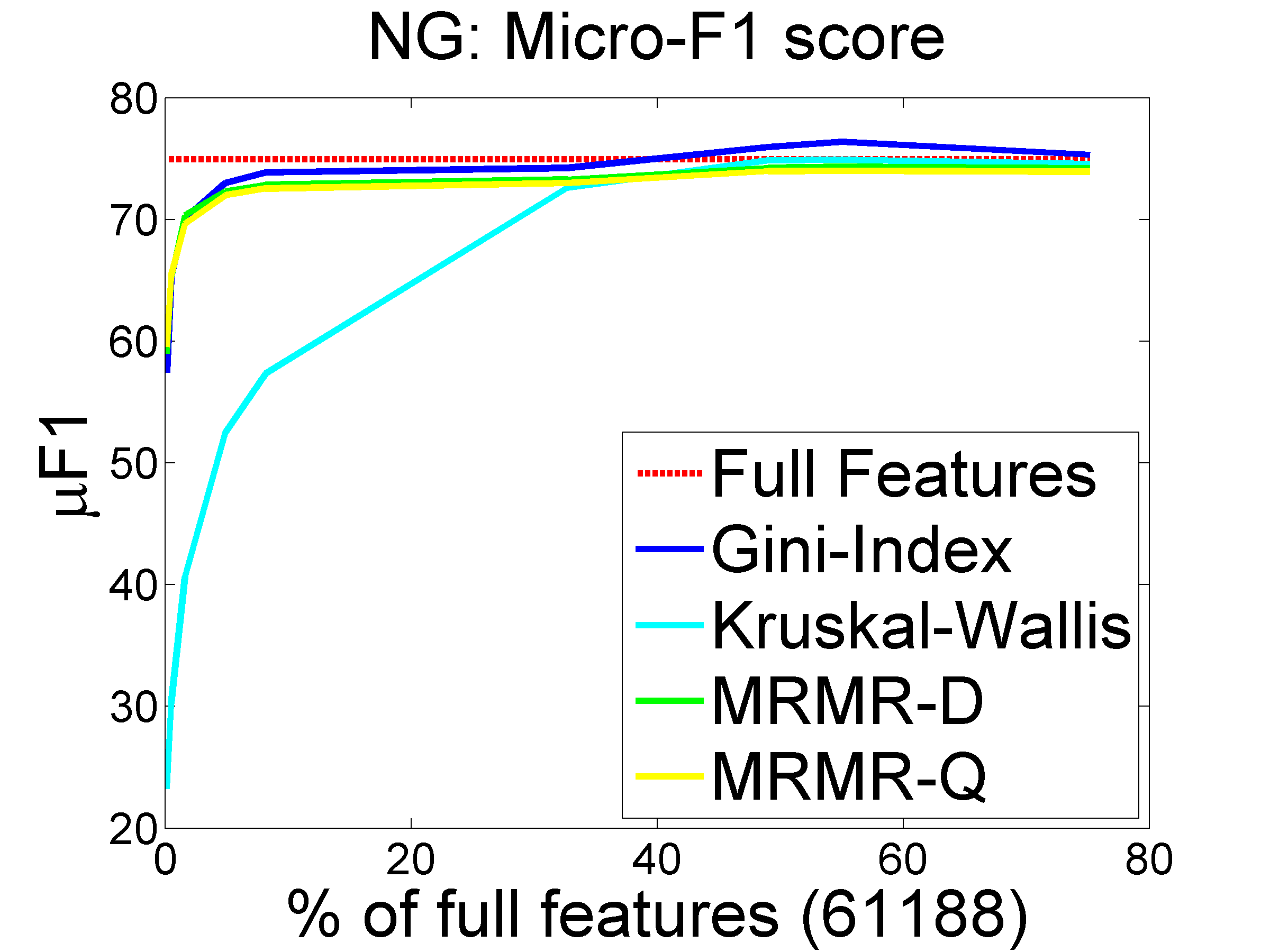}%
            \label{fig:ngsub1}}
        \hfil
        \subfloat[]{
            \includegraphics[width=.225\linewidth,height=4cm]{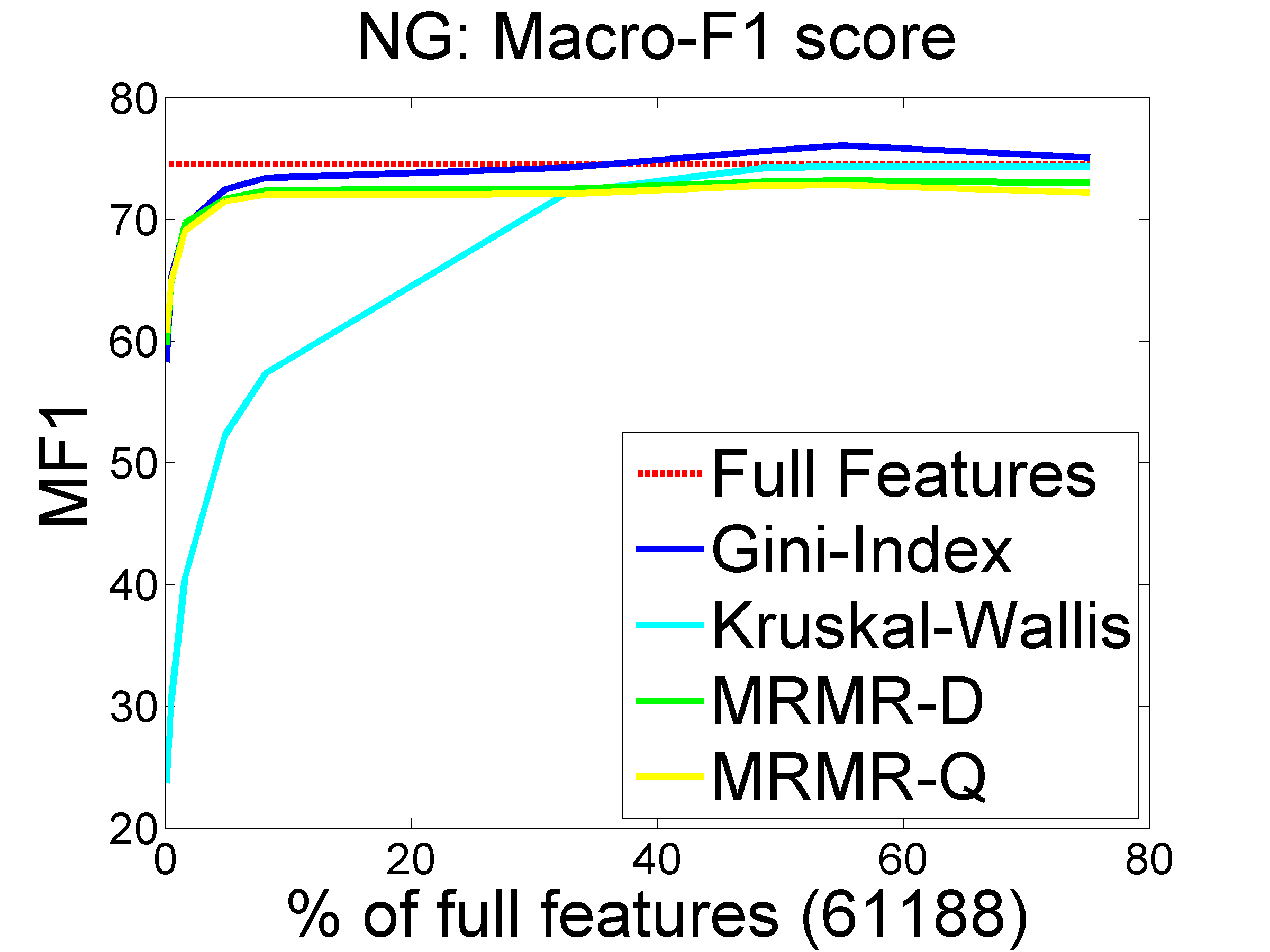}%
            \label{fig:ngsub2}}
        \hfil
        \subfloat[]{
            \includegraphics[width=.225\linewidth,height=4cm]{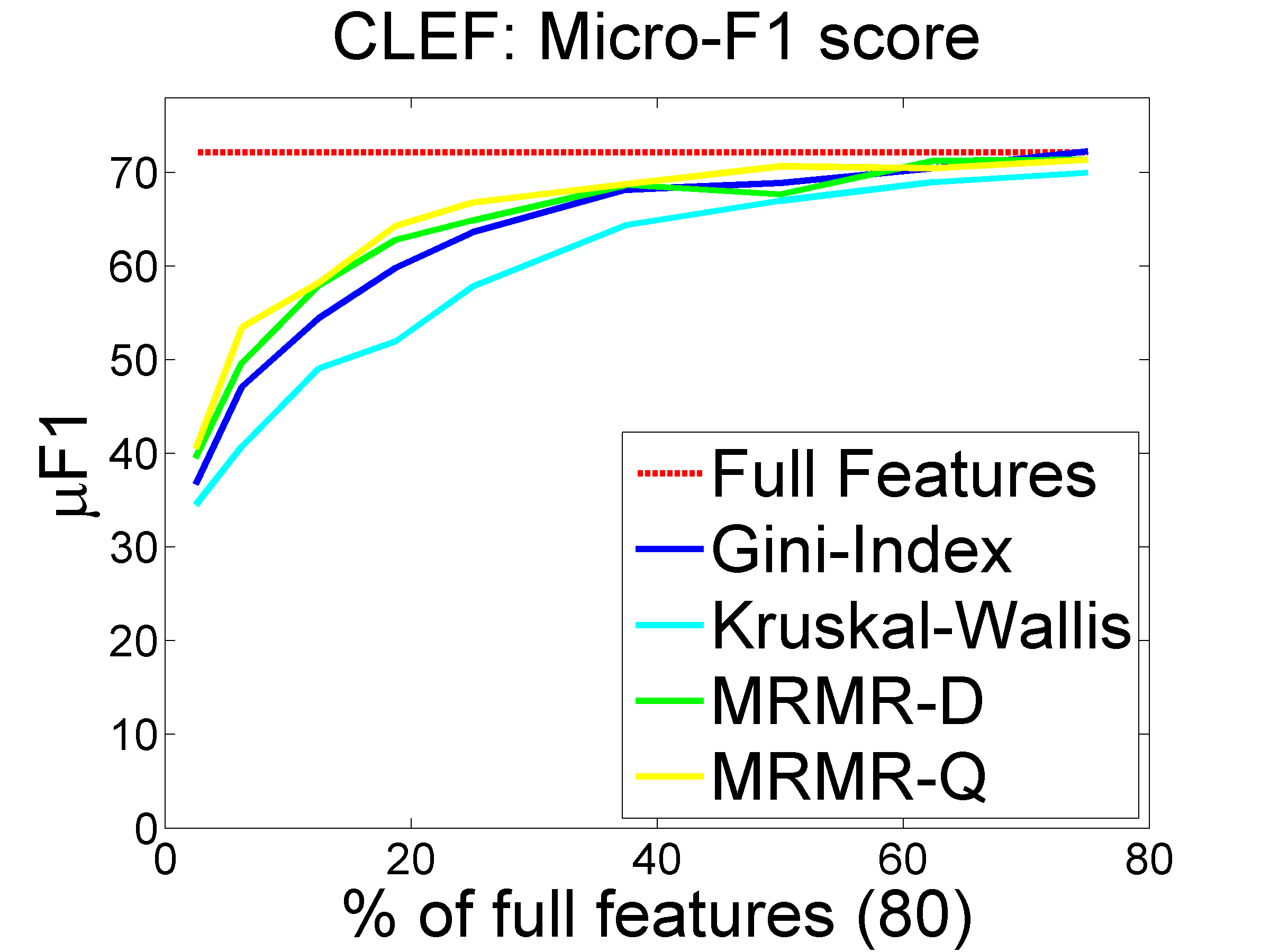}%
            \label{fig:clefsub1}}
        \hfil
        \subfloat[]{
            \includegraphics[width=.225\linewidth,height=4cm]{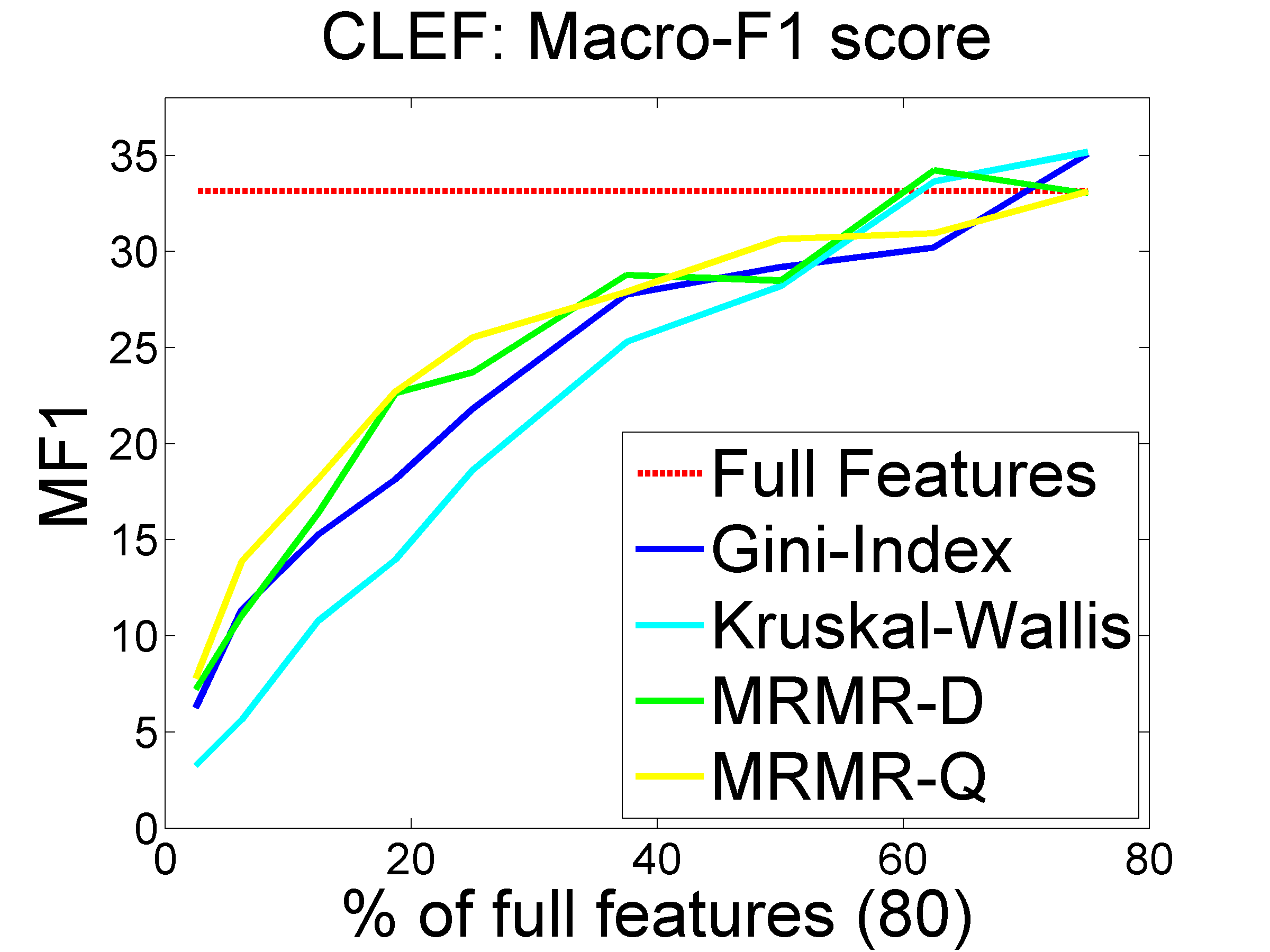}%
            \label{fig:clefsub2}}%
        \\
        \subfloat[]{
            \includegraphics[width=.225\linewidth,height=4cm]{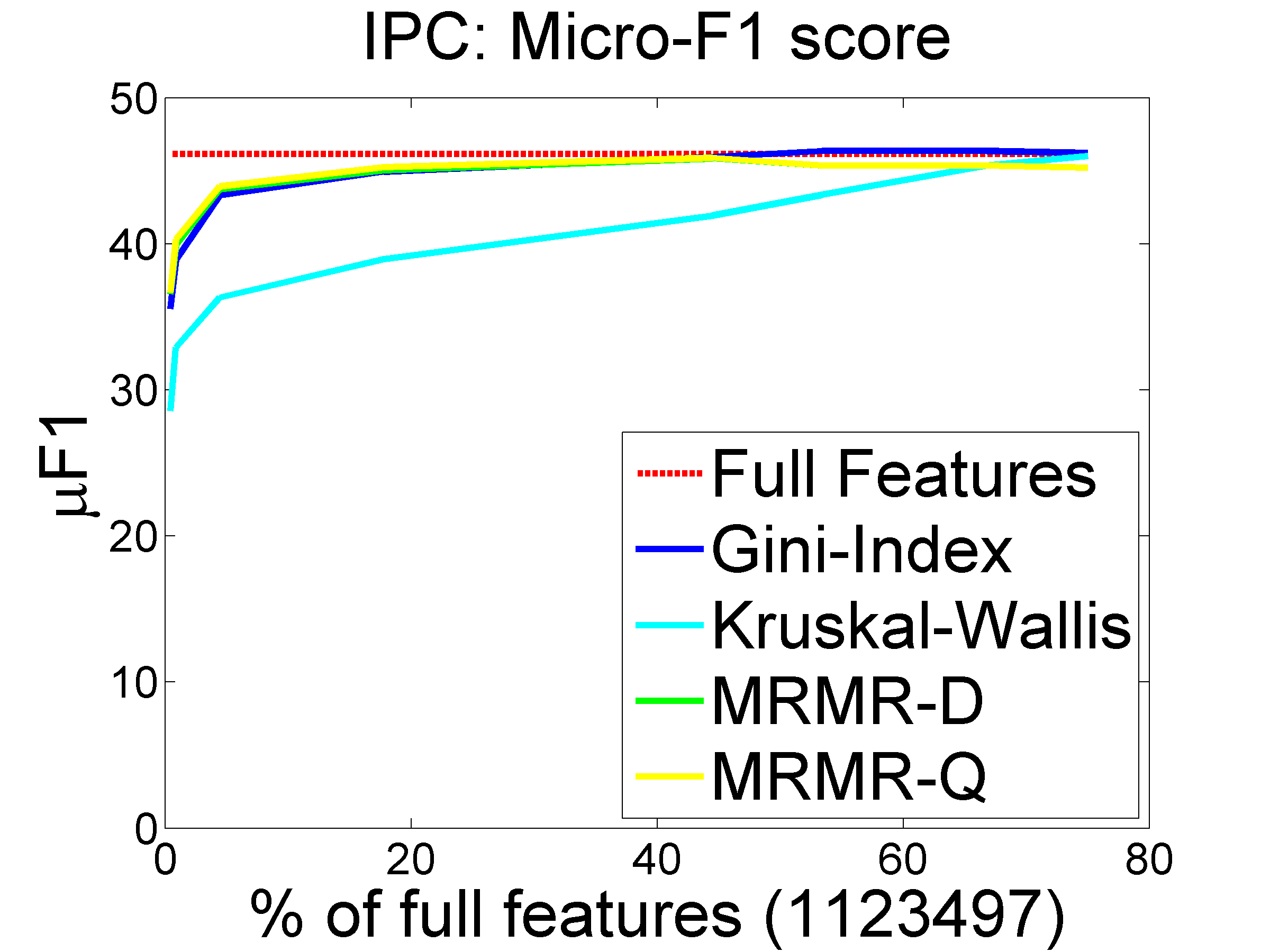}%
            \label{fig:ipcsub1}}%
        \hfil
        \subfloat[]{
            \includegraphics[width=.225\linewidth,height=4cm]{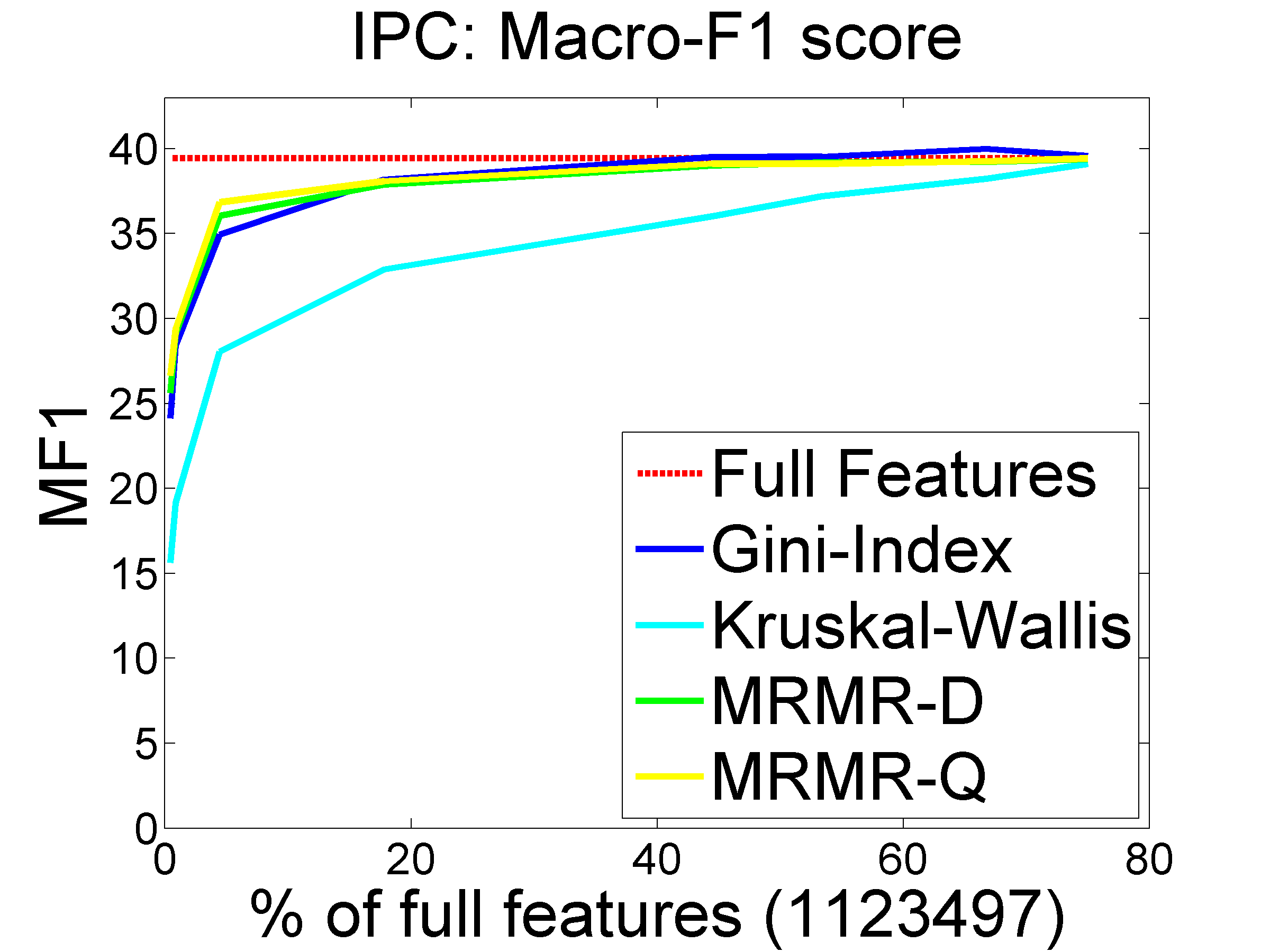}%
            \label{fig:ipcsub2}}%
        \hfil
        \subfloat[]{
            \includegraphics[width=.225\linewidth,height=4cm]{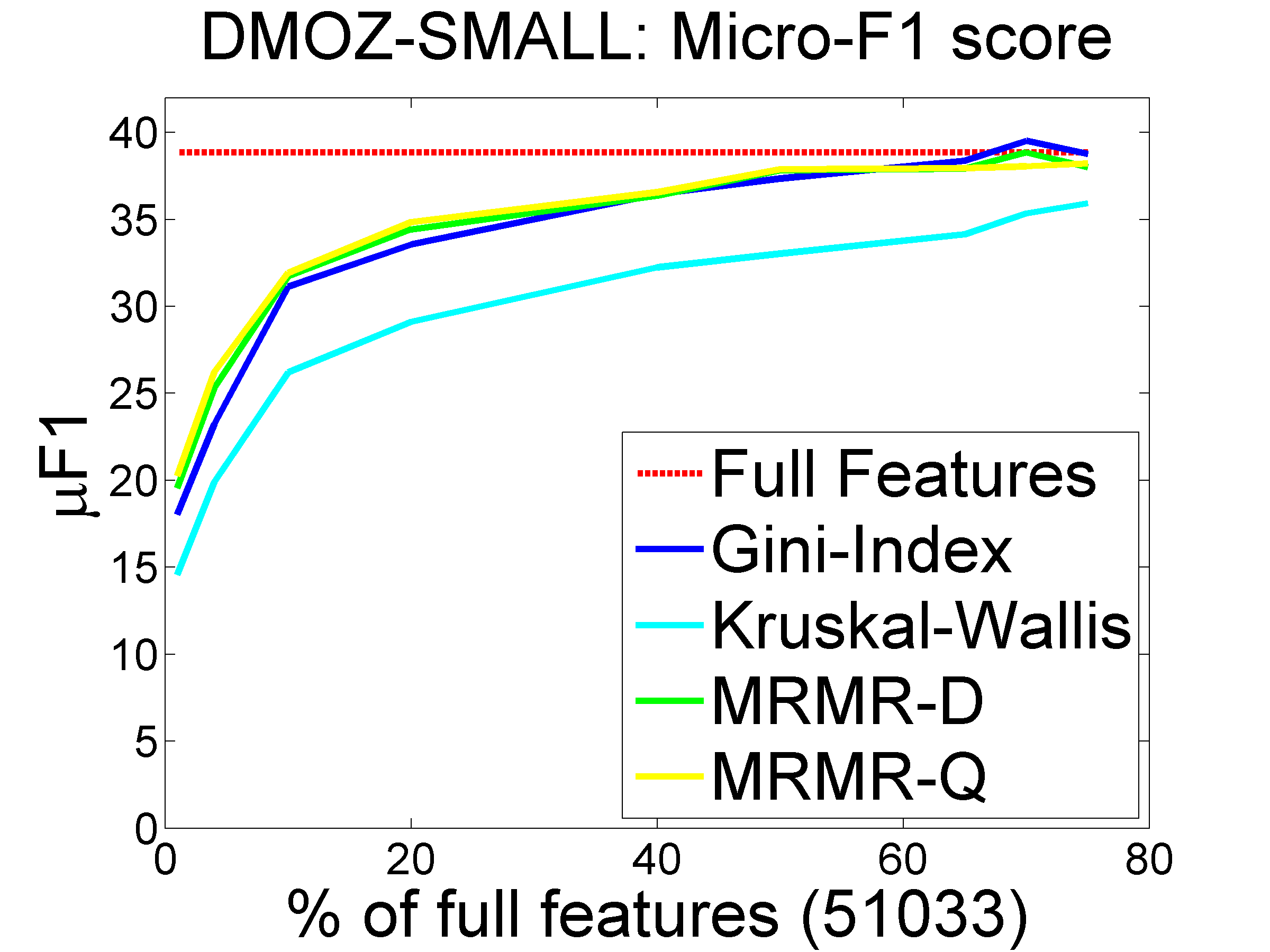}%
            \label{fig:dmozsmallsub1}}%
        \hfil
        \subfloat[]{
            \includegraphics[width=.225\linewidth,height=4cm]{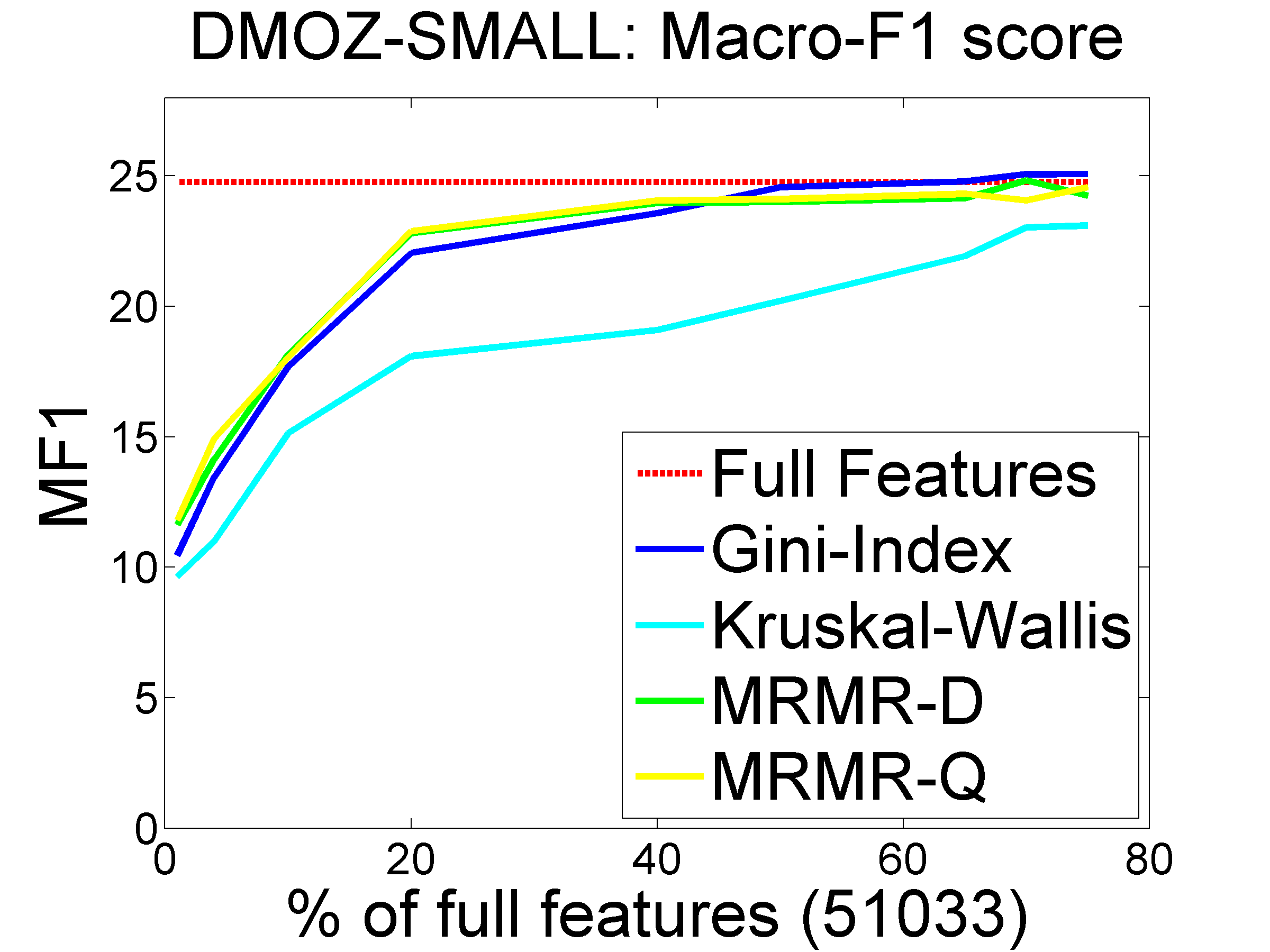}%
            \label{fig:dmozsmallsub2}}%
        \\
        \subfloat[]{
            \includegraphics[width=.225\linewidth,height=4cm]{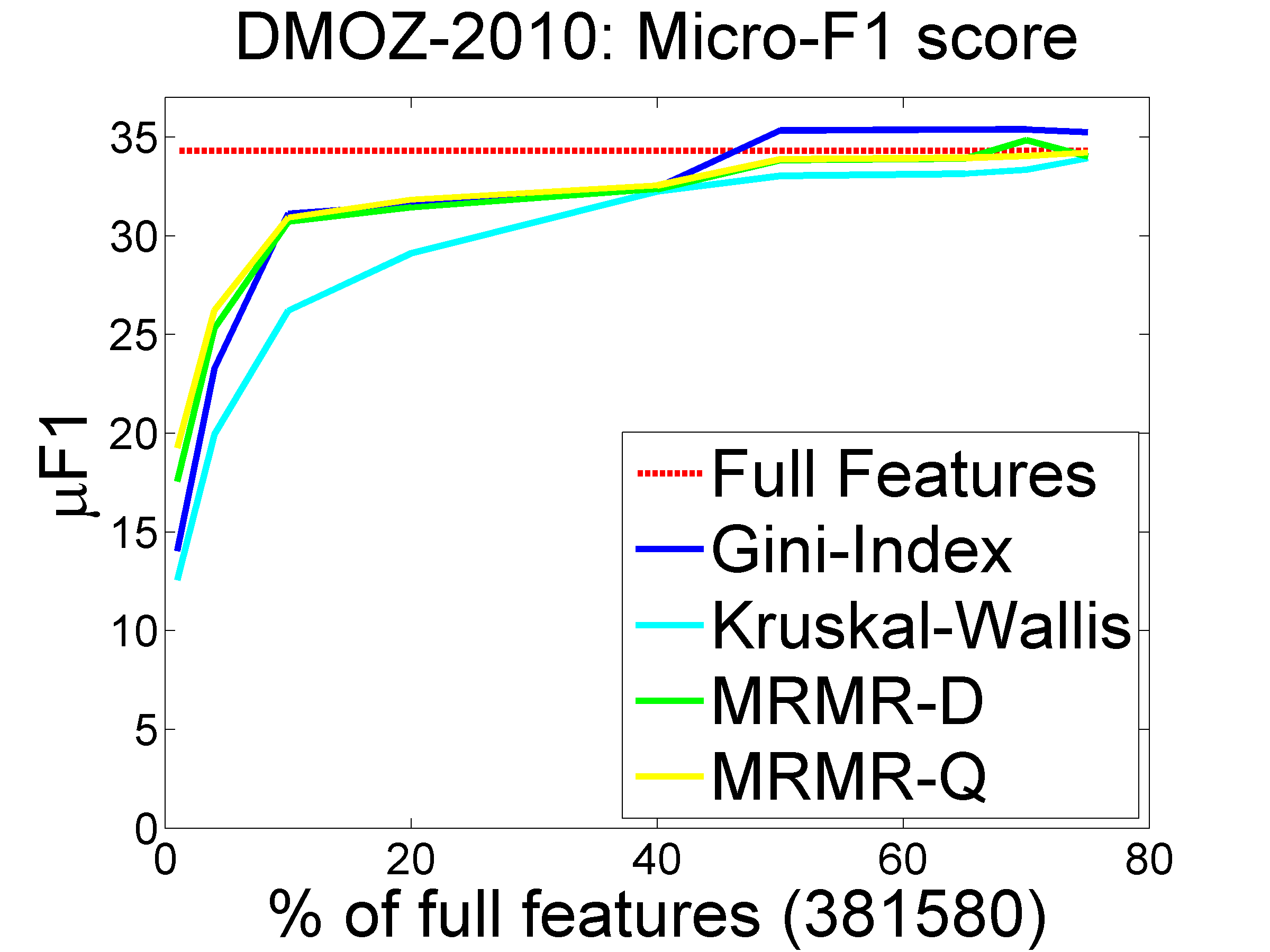}%
            \label{fig:dmoz2010sub1}}%
        \hfil
        \subfloat[]{
            \includegraphics[width=.225\linewidth,height=4cm]{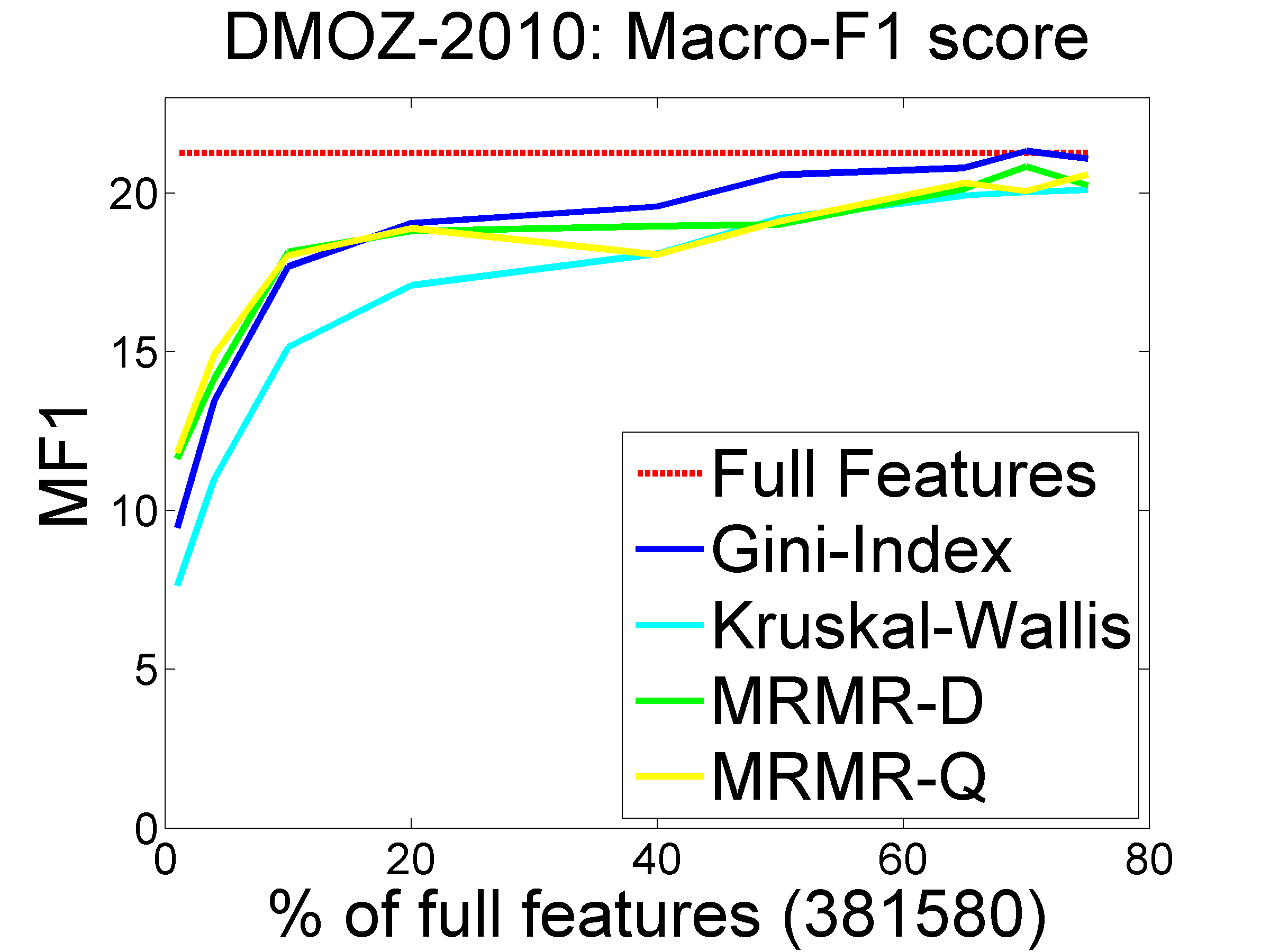}%
            \label{fig:dmoz2010sub2}}%
        \hfil
        \subfloat[]{
            \includegraphics[width=.225\linewidth,height=4cm]{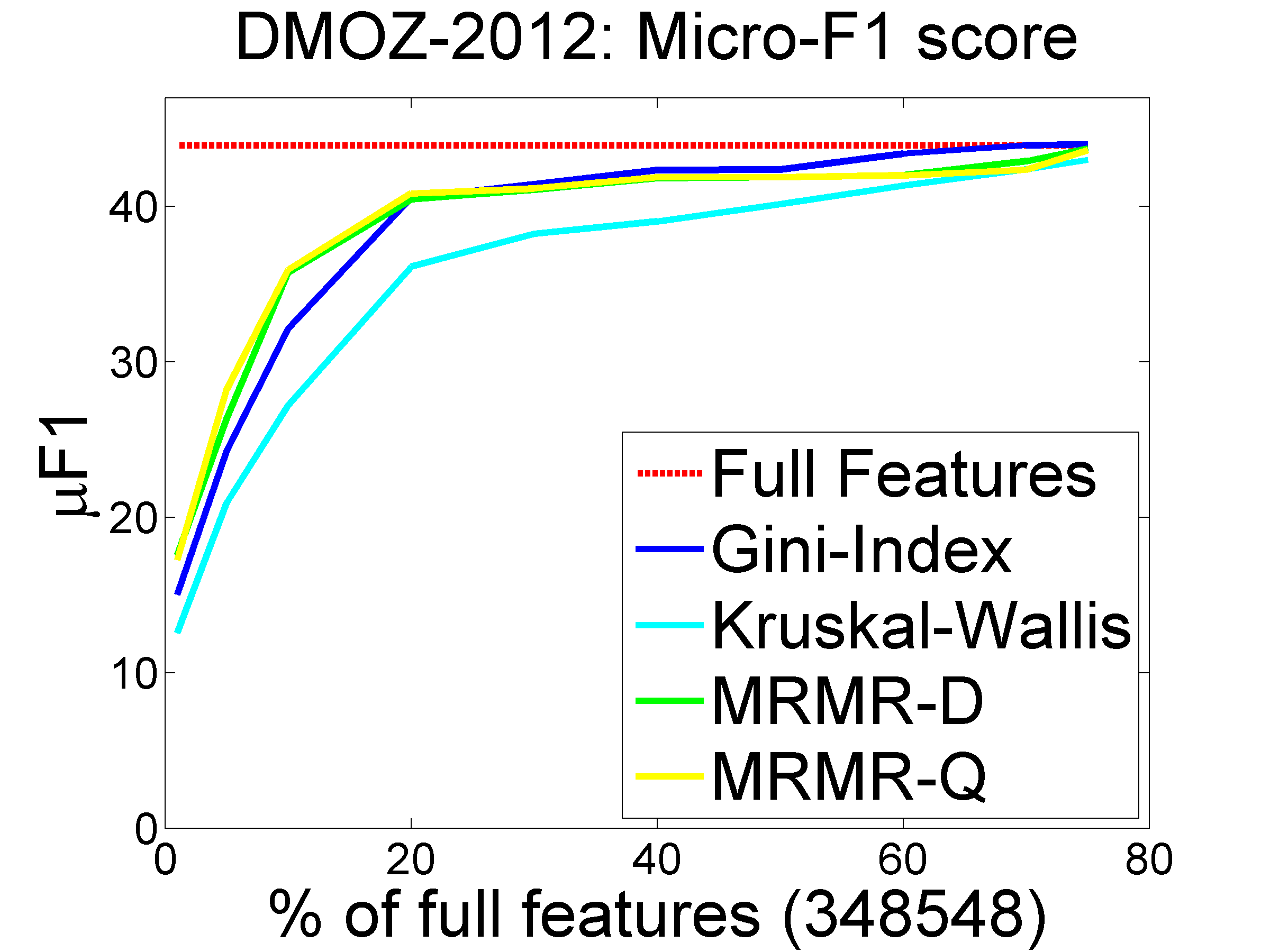}%
            \label{fig:dmoz2012sub1}}%
        \hfil
        \subfloat[]{
            \includegraphics[width=.225\linewidth,height=4cm]{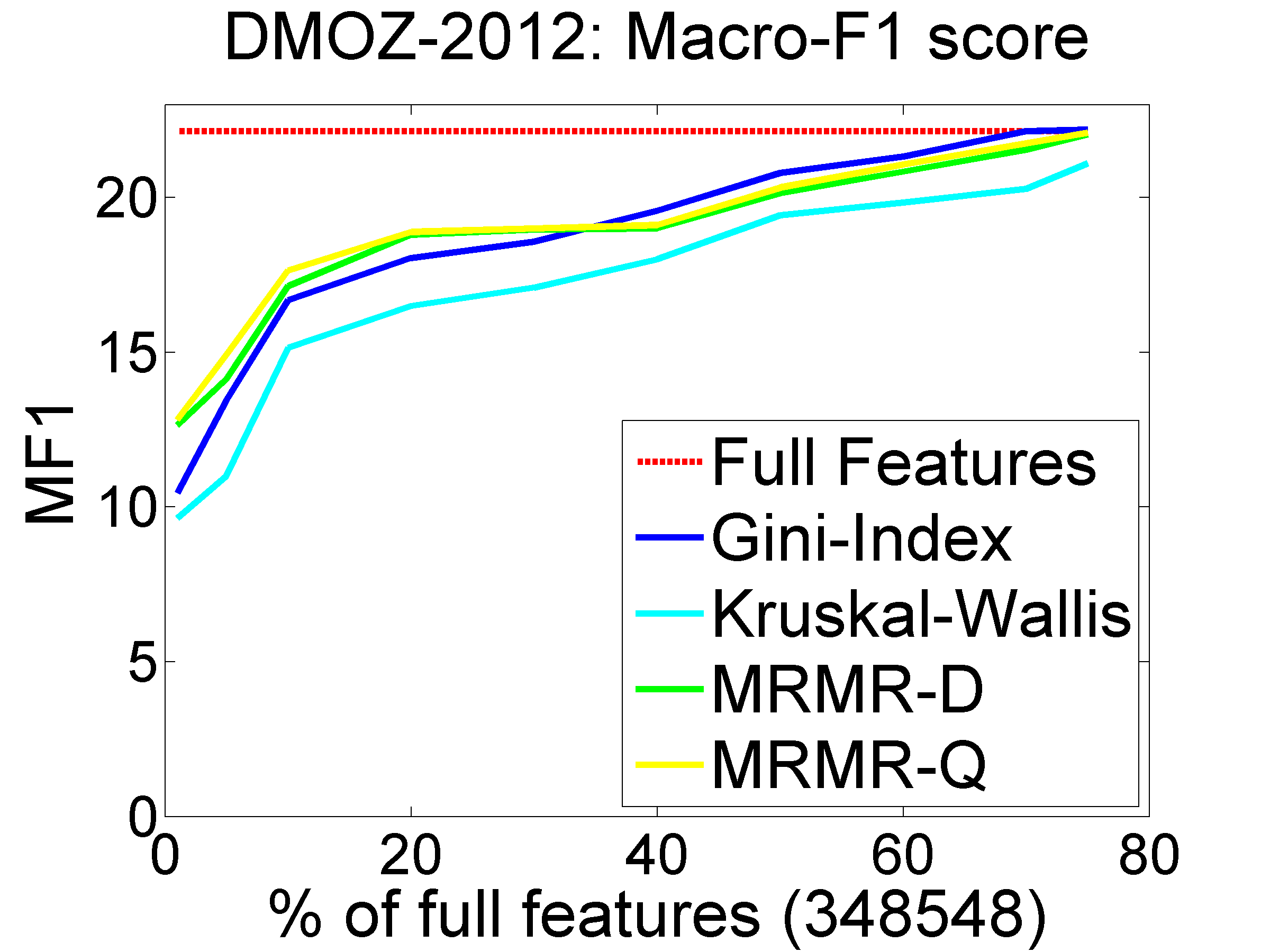}%
            \label{fig:dmoz2012sub2}}%
        \caption{\textbf{Performance comparison of LR + $l_1$-norm models with varying percentage (\%) of  features selected using different feature selection (global) methods on text and image datasets.}}
    \label{figaccuracyL1}
\end{figure*}

\begin{figure*}
       \center
        \subfloat[]{
            \includegraphics[width=.225\linewidth,height=4cm]{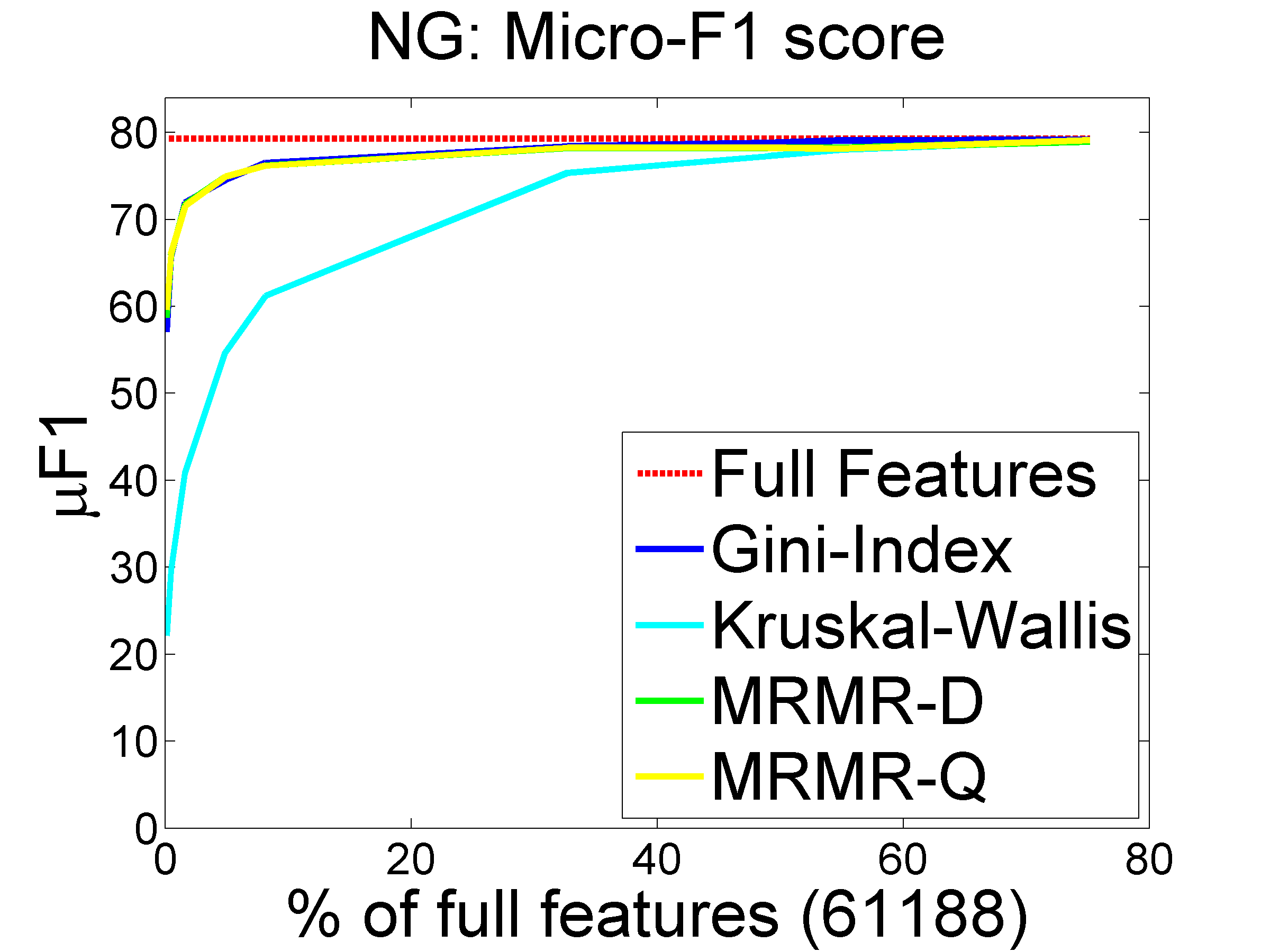}%
            \label{fig:ngsub1L2}}%
        \hfil
        \subfloat[]{
            \includegraphics[width=.225\linewidth,height=4cm]{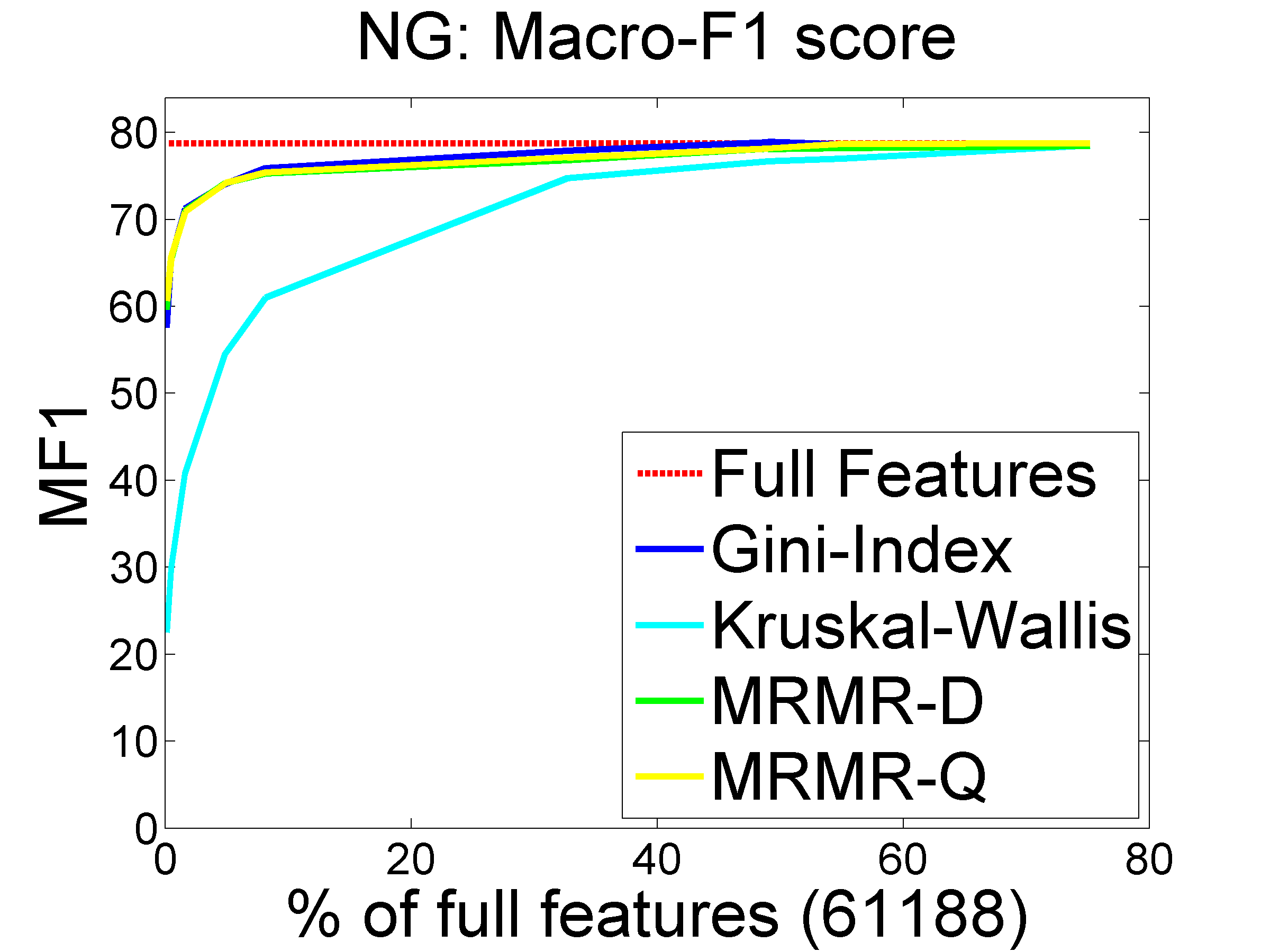}%
            \label{fig:ngsub2L2}}%
        \hfil
        \subfloat[]{
            \includegraphics[width=.225\linewidth,height=4cm]{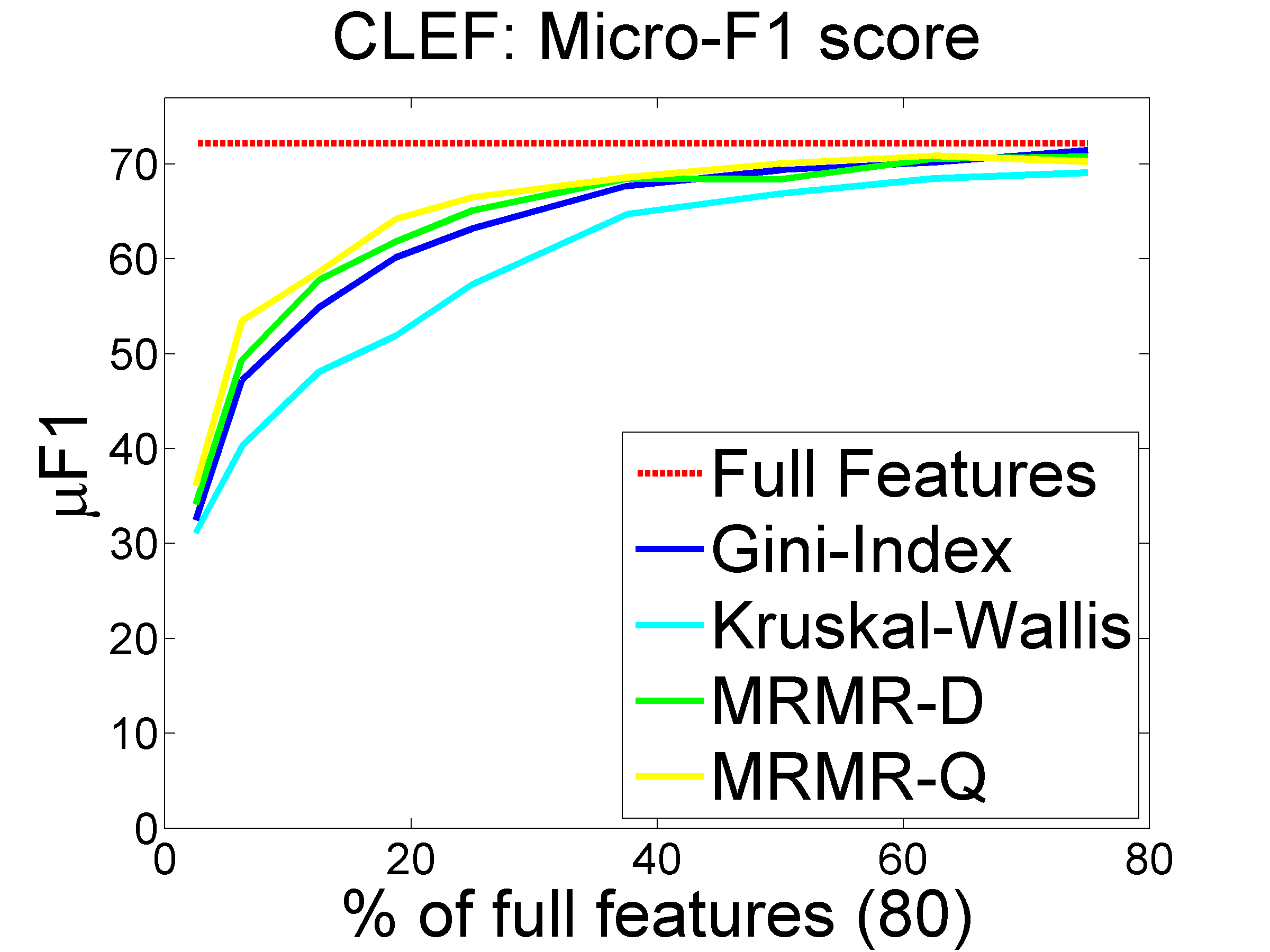}%
            \label{fig:clefsub1L2}}%
        \hfil
        \subfloat[]{
            \includegraphics[width=.225\linewidth,height=4cm]{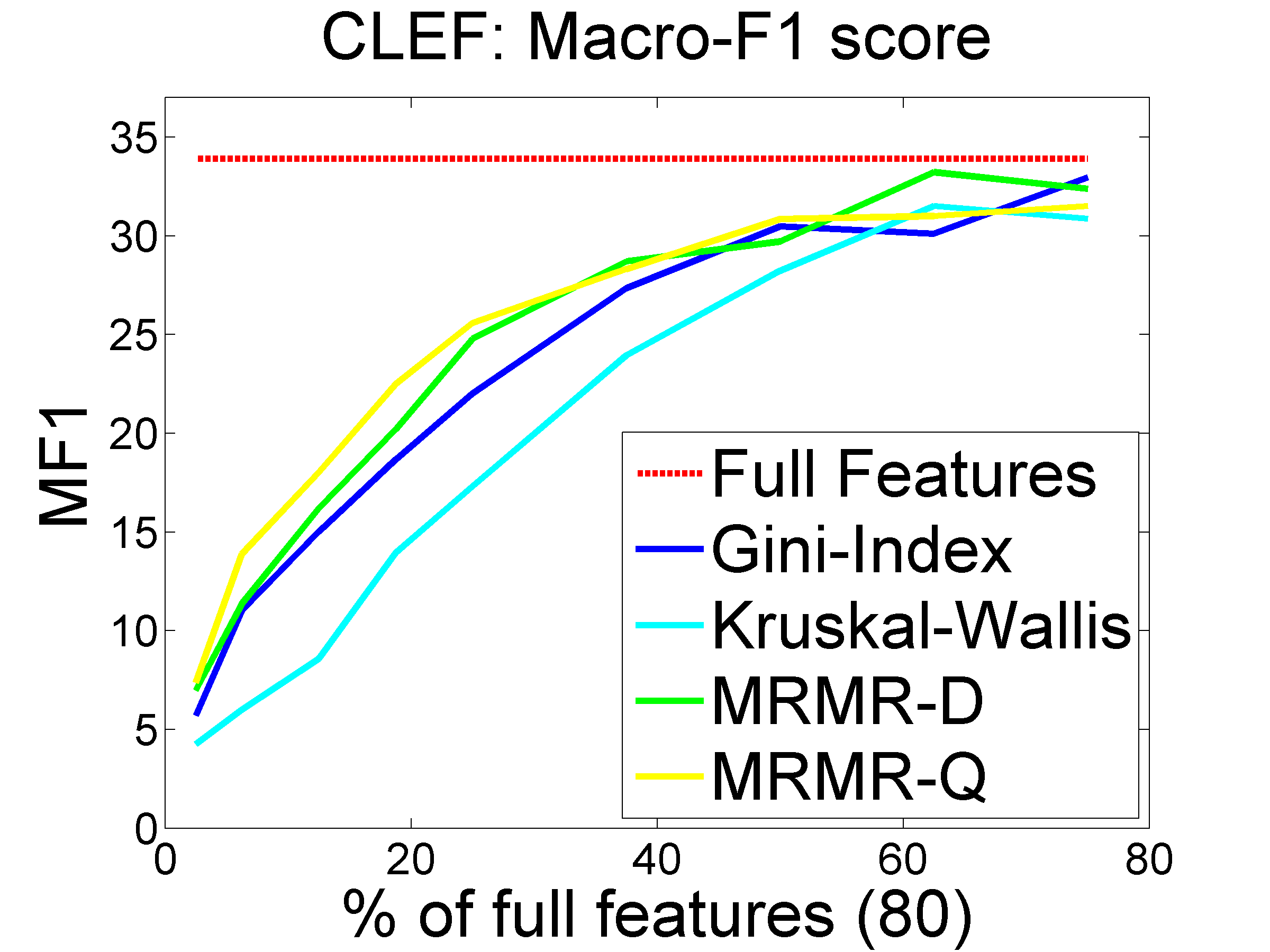}%
            \label{fig:clefsub2L2}}%
        \\
        \subfloat[]{
            \includegraphics[width=.225\linewidth,height=4cm]{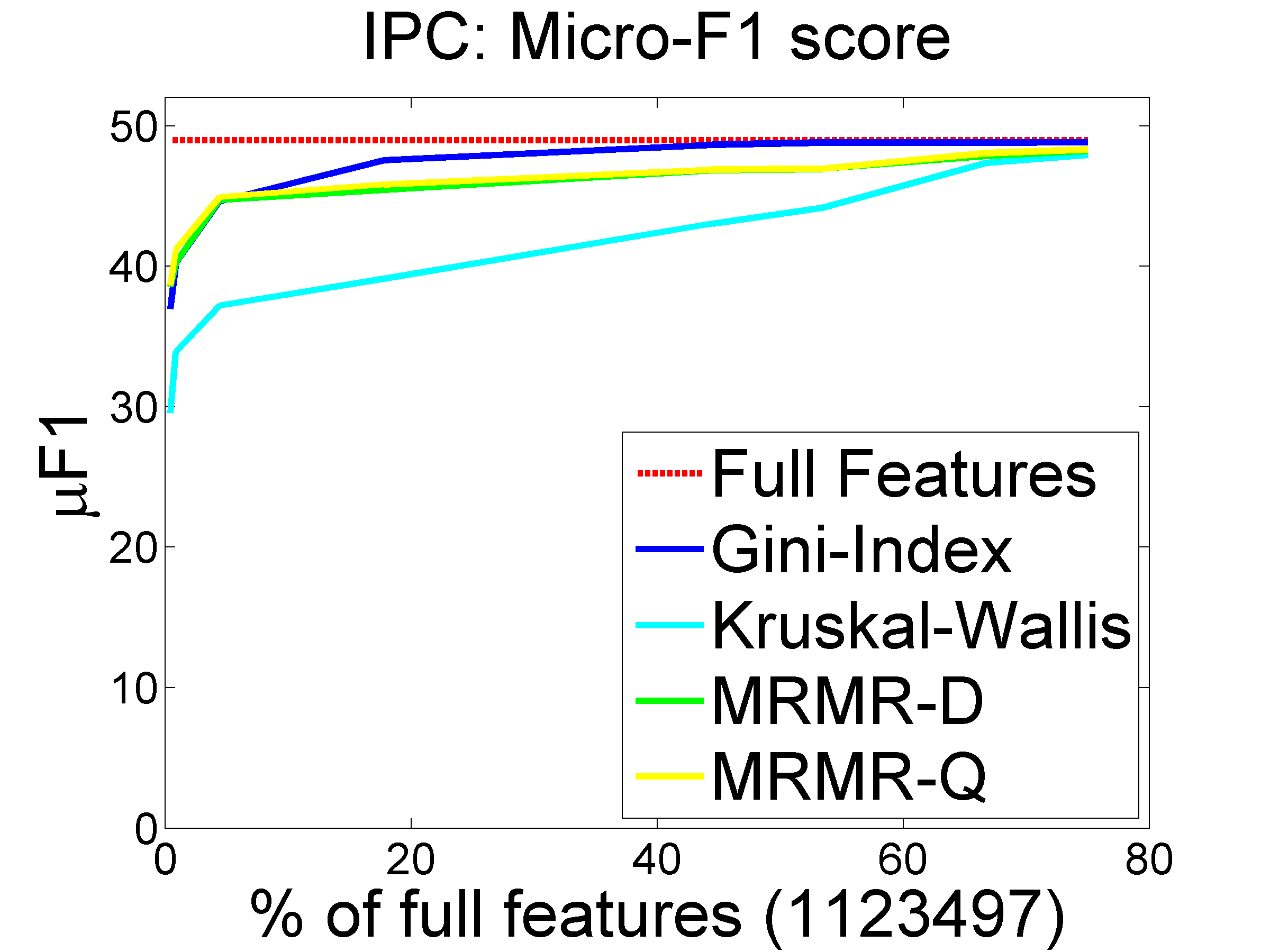}%
            \label{fig:ipcsub1L2}}%
        \hfil
        \subfloat[]{
            \includegraphics[width=.225\linewidth,height=4cm]{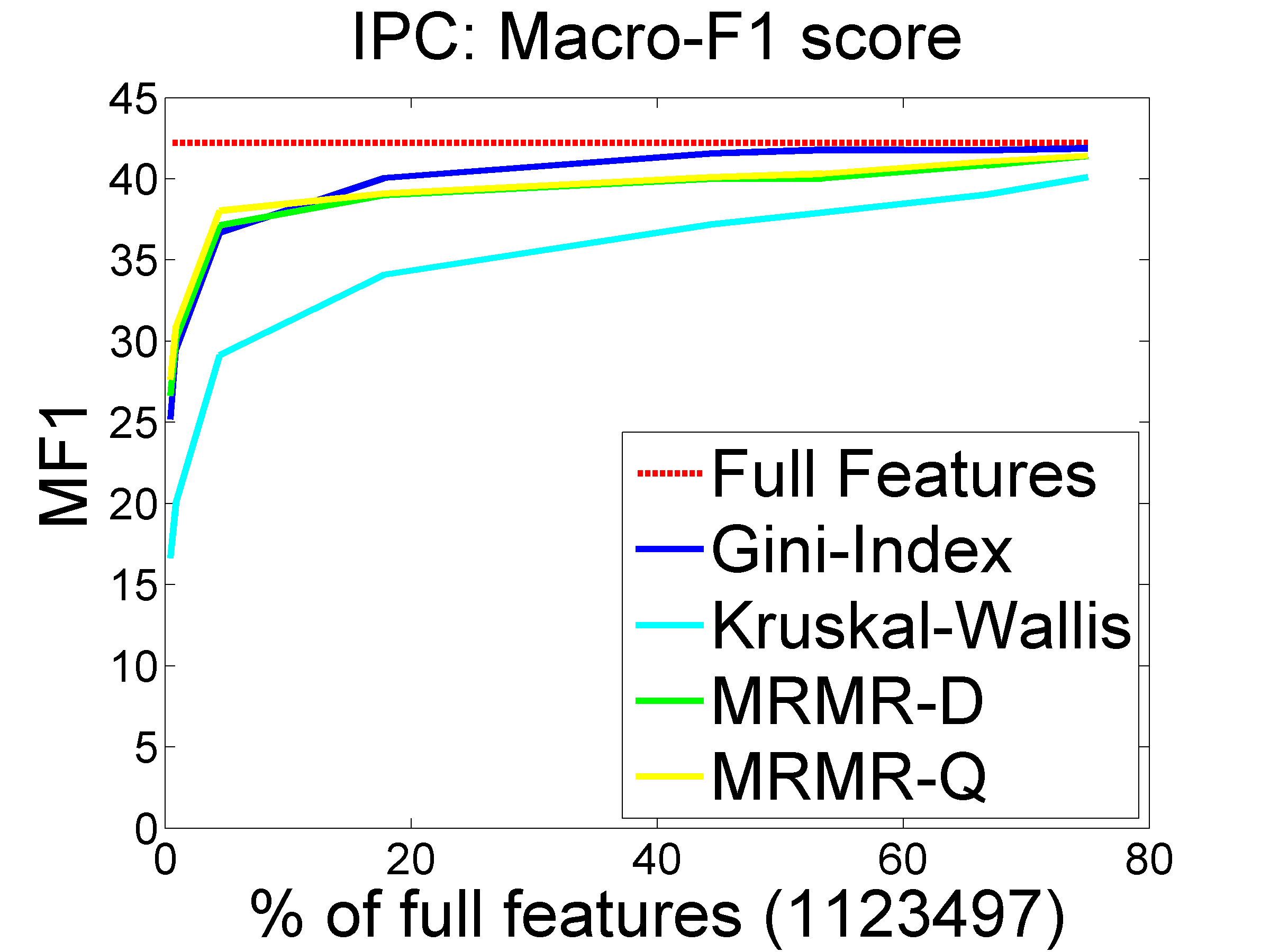}%
            \label{fig:ipcsub2L2}}%
        \hfil
        \subfloat[]{
            \includegraphics[width=.225\linewidth,height=4cm]{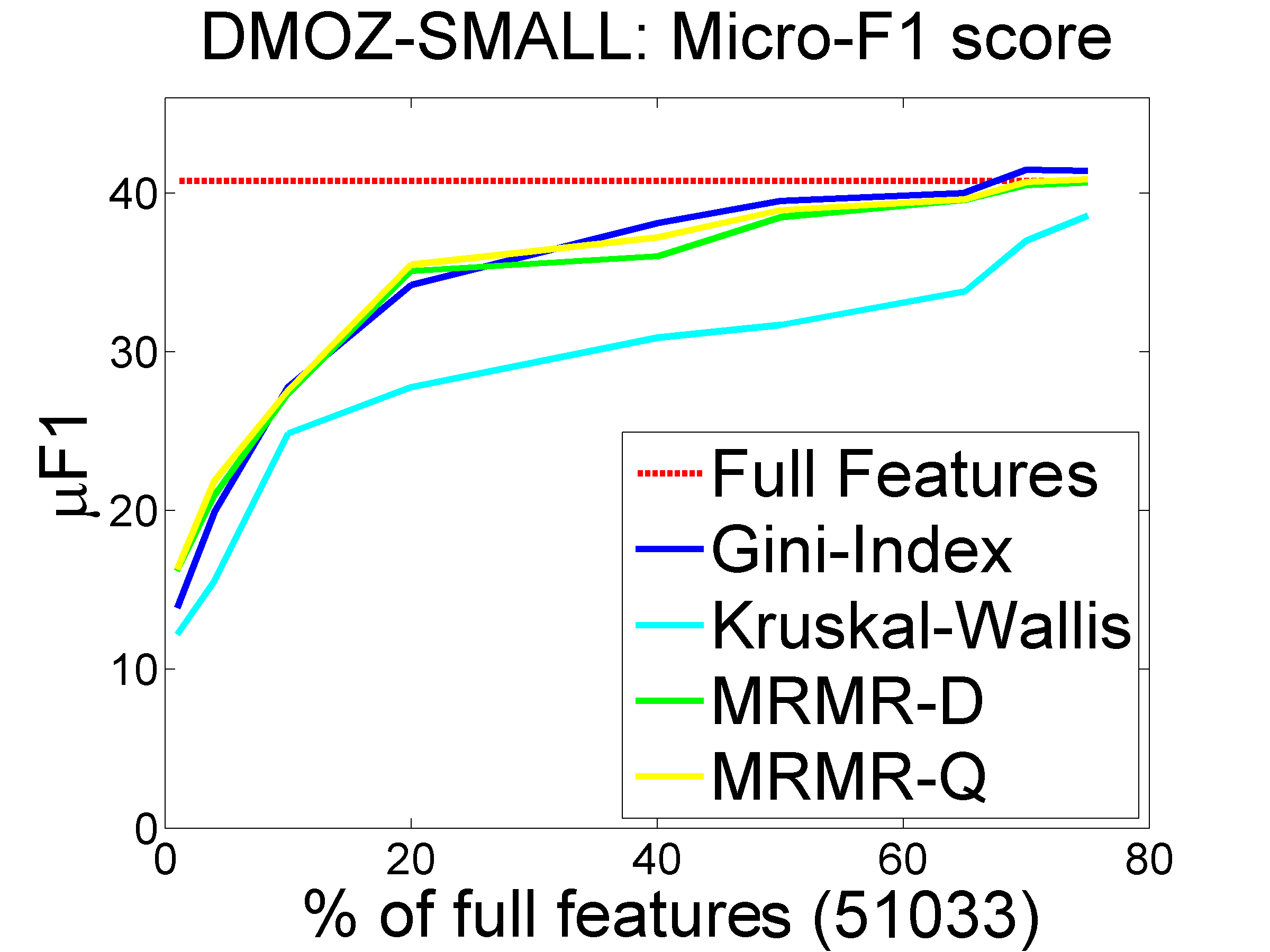}%
            \label{fig:dmozsmallsub1L2}}%
        \hfil
        \subfloat[]{
            \includegraphics[width=.225\linewidth,height=4cm]{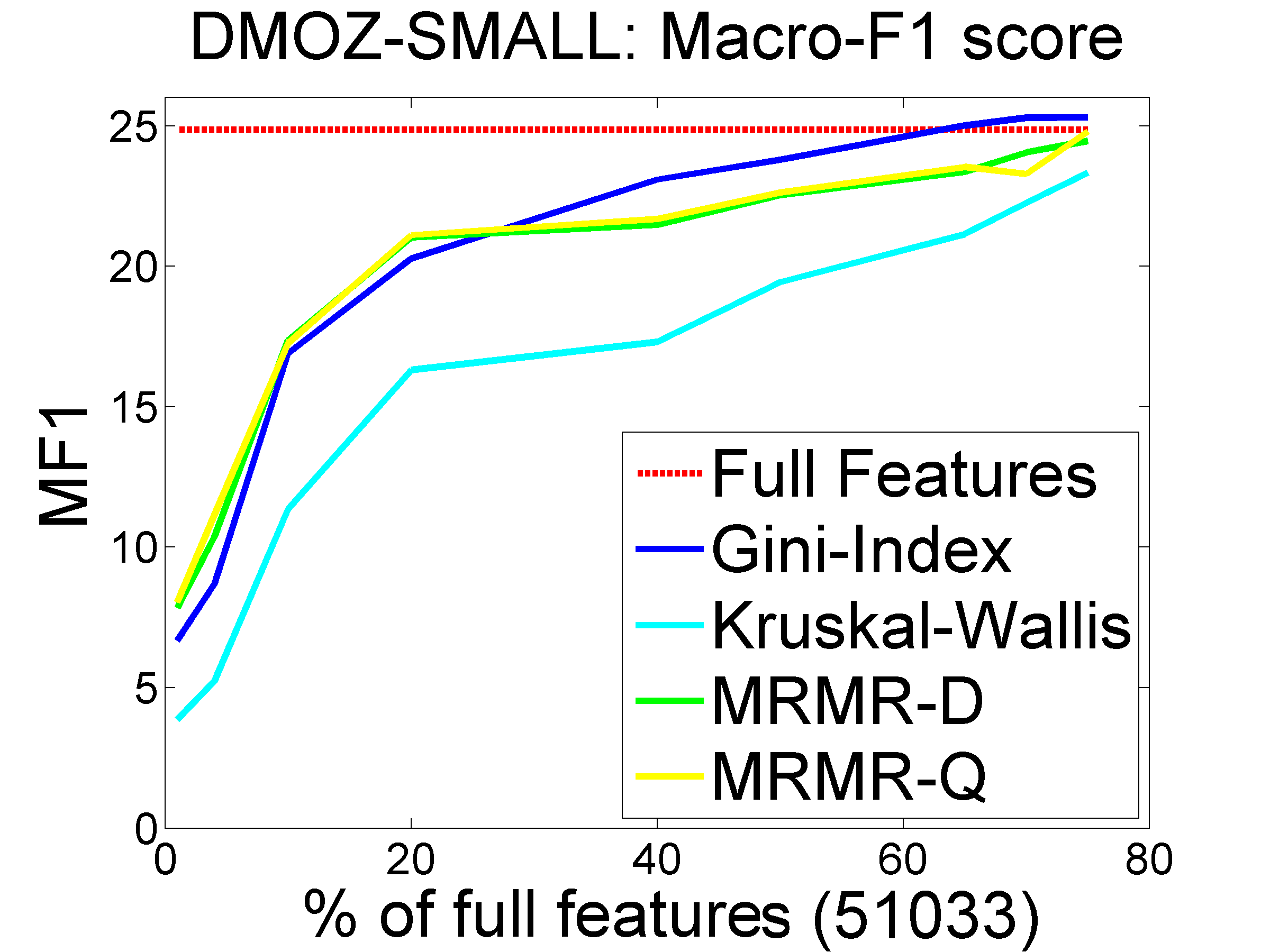}%
            \label{fig:dmozsmallsub2L2}}%
        \\
        \subfloat[]{
            \includegraphics[width=.225\linewidth,height=4cm]{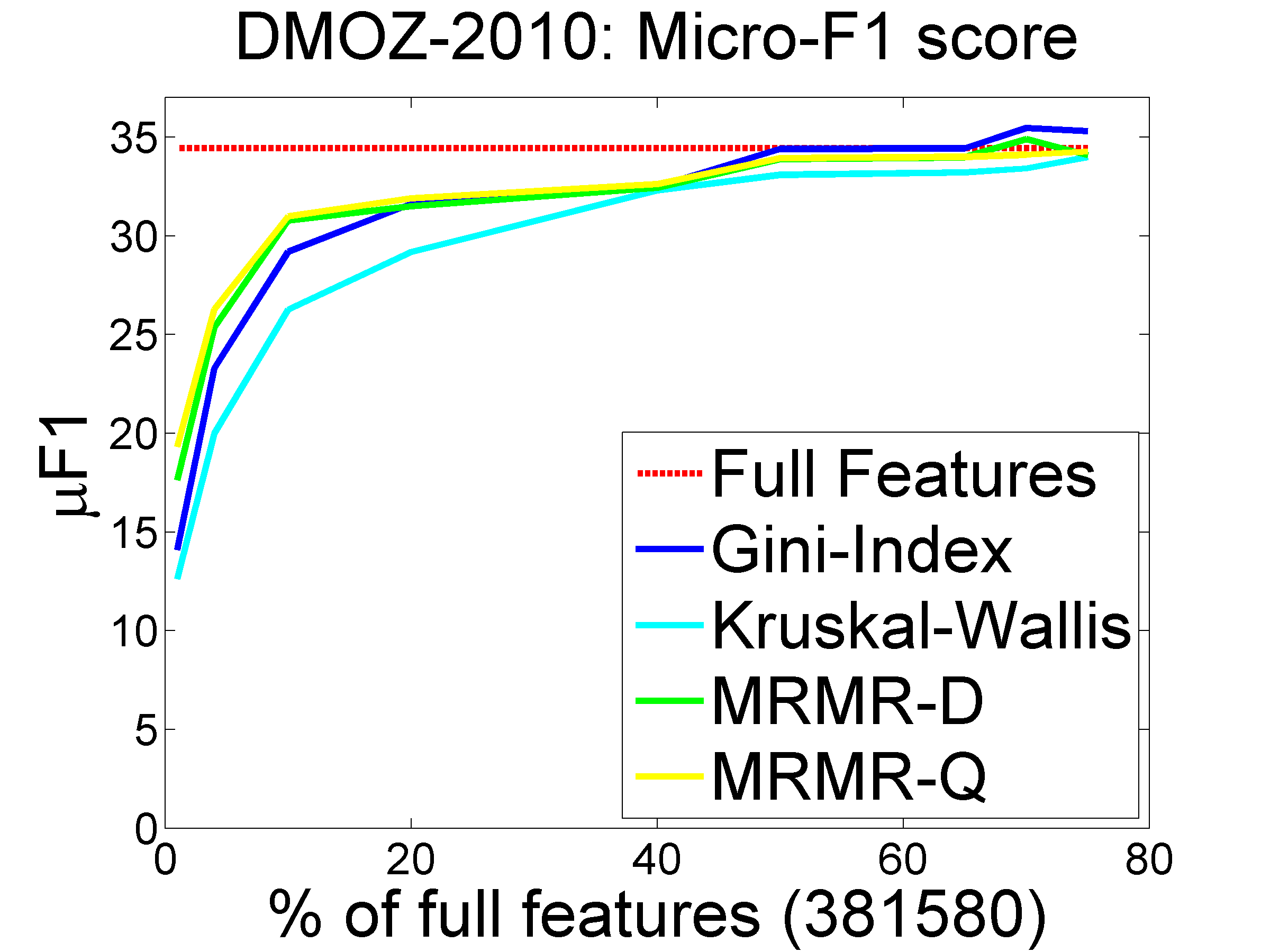}%
            \label{fig:dmoz2010sub1L2}}%
        \hfil
        \subfloat[]{
            \includegraphics[width=.225\linewidth,height=4cm]{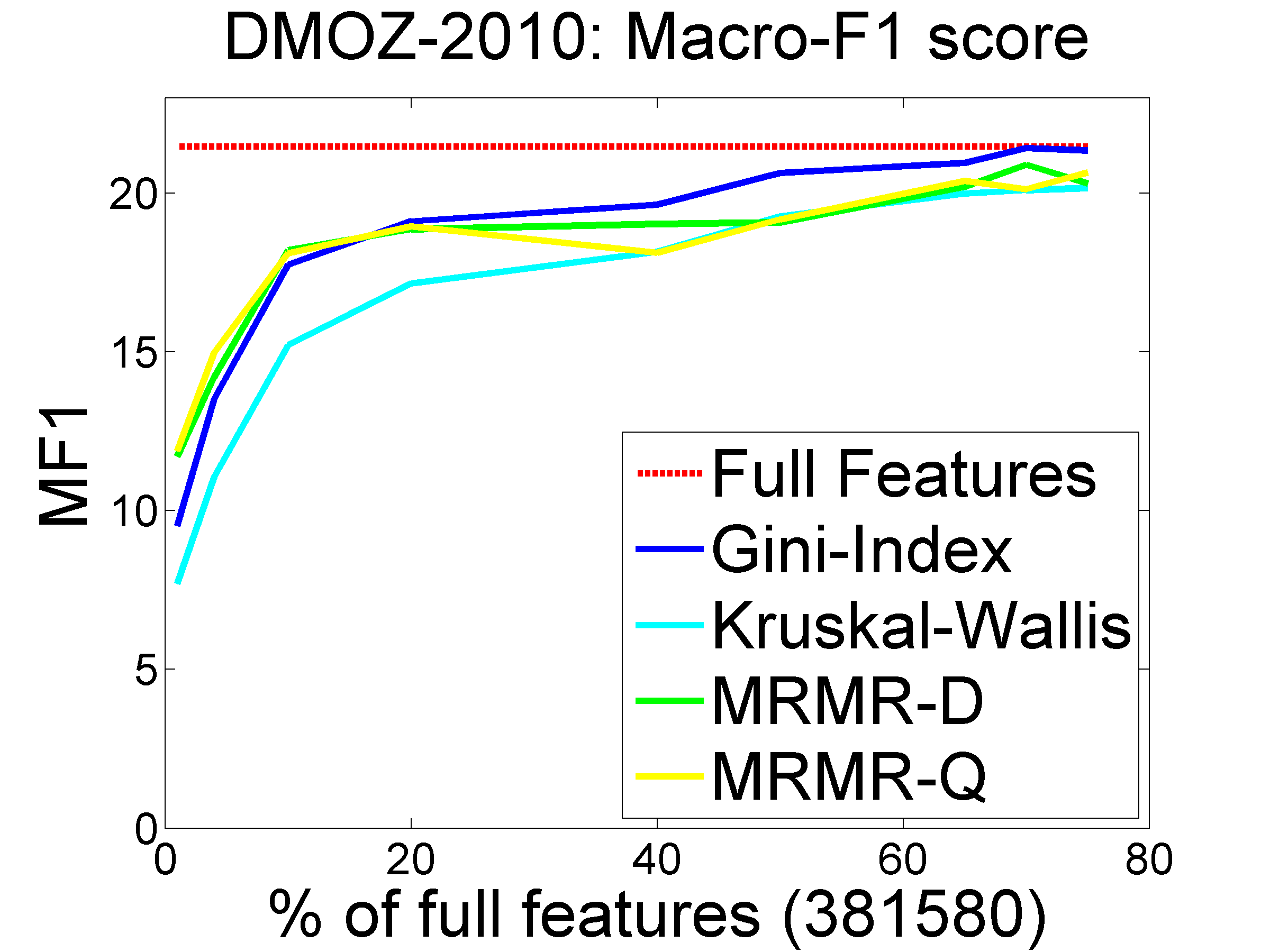}%
            \label{fig:dmoz2010sub2L2}}%
        \hfil
        \subfloat[]{
            \includegraphics[width=.225\linewidth,height=4cm]{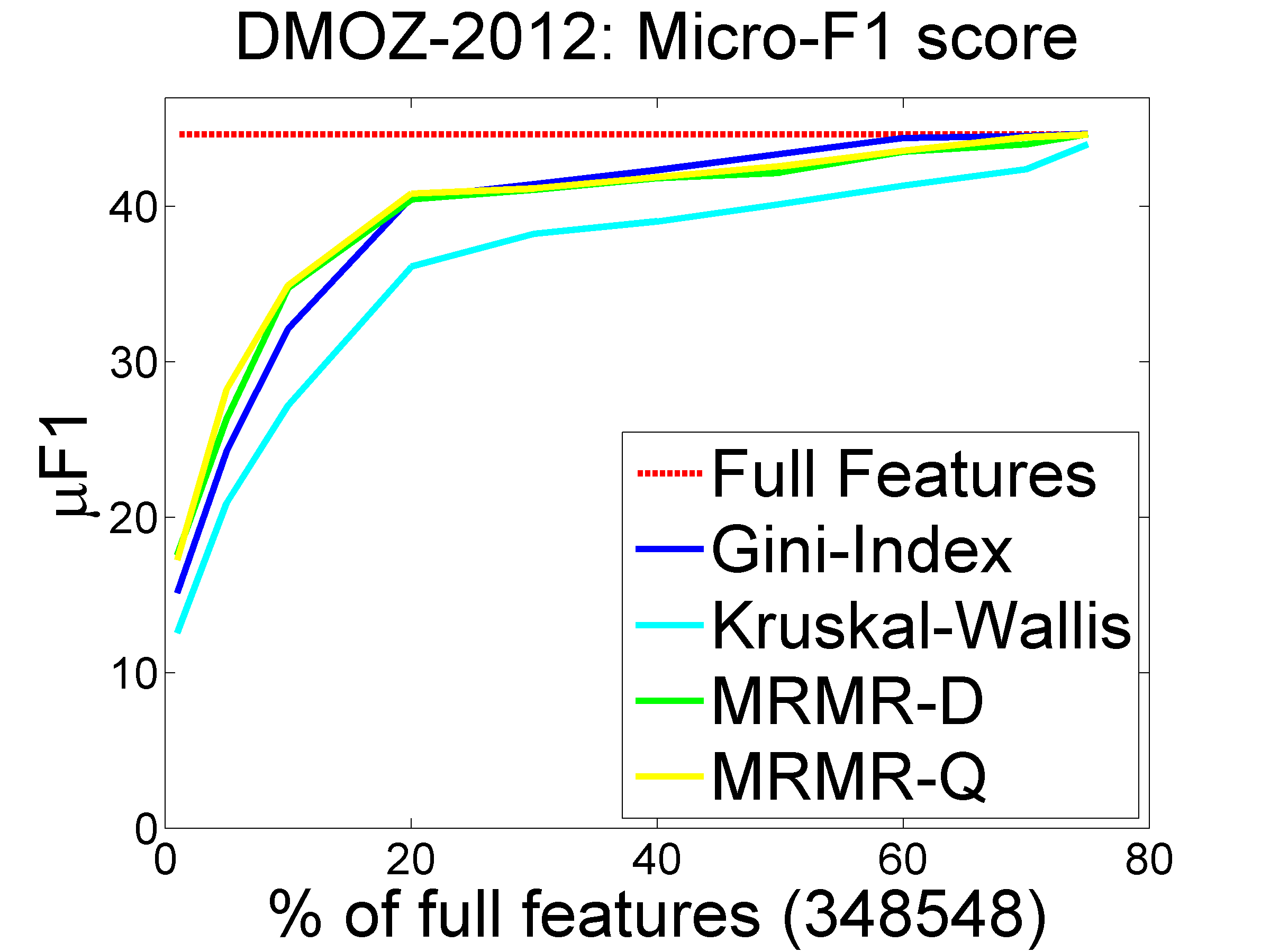}%
            \label{fig:dmoz2012sub1L2}}%
        \hfil
        \subfloat[]{
            \includegraphics[width=.225\linewidth,height=4cm]{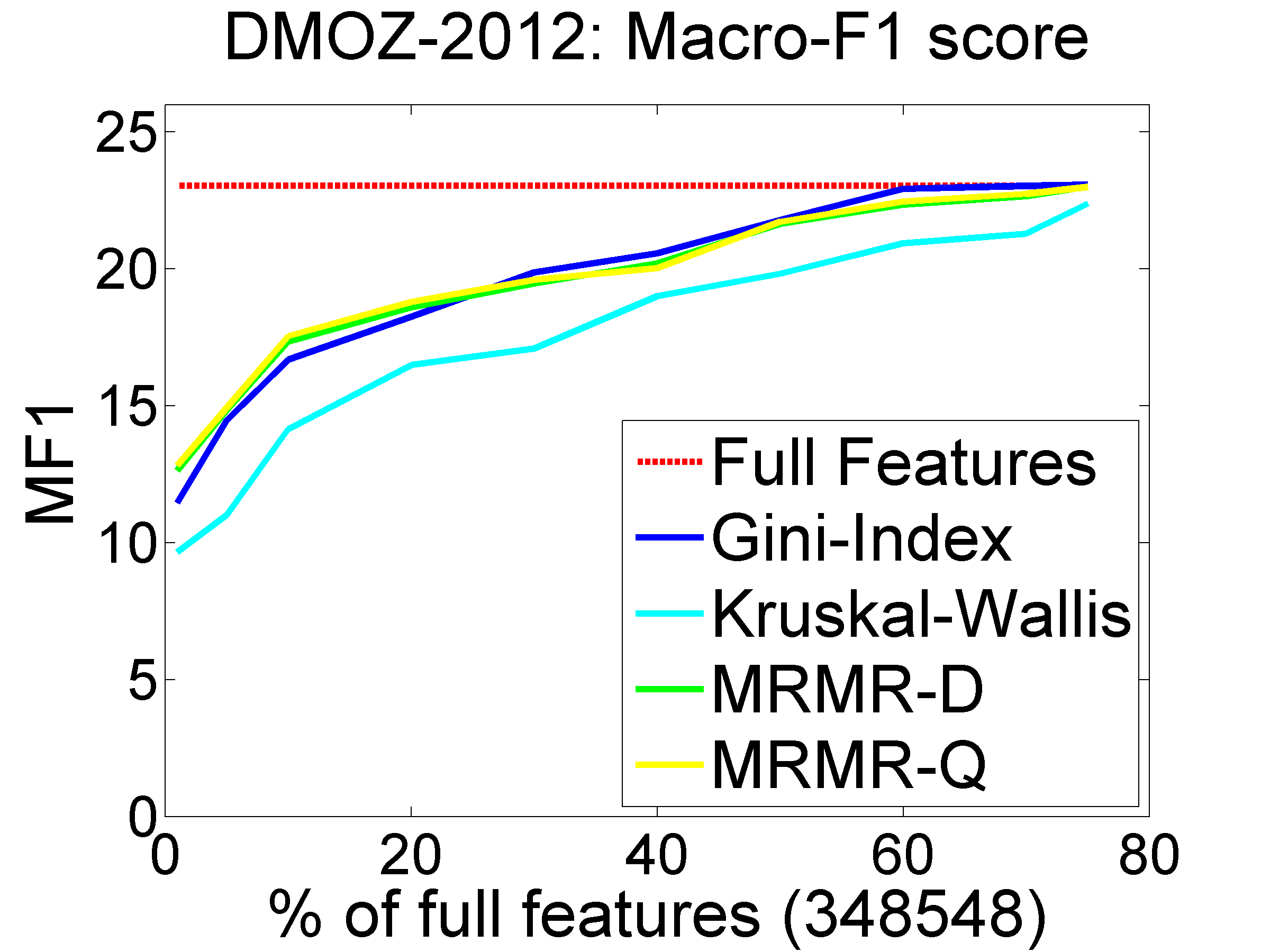}%
            \label{fig:dmoz2012sub2L2}}%
        \caption{\textbf{Performance comparison of LR + $l_2$-norm models with varying percentage (\%) of  features selected using different feature selection (global) methods on text and image datasets.}}
    \label{figaccuracyL2}
\end{figure*}

\subsection{Experimental Details}
For all the experiments, we divide the training dataset into train and small validation dataset in the ratio 90:10. The train dataset is used to train TD classifiers whereas the validation dataset is used to tune the parameter. The model is trained for a range of mis-classification penalty parameter ($\lambda$) values in the set \big\{0.001, 0.01, 0.1, 1, 10, 100, 1000\big\} with best value selected using a validation dataset. Adopting the best parameter, we retrain the models on the entire training dataset and measure the performance on a separate held-out test dataset. For feature selection, we choose the best set of features using the validation dataset by varying the number of features between 1$\%$ and 75$\%$ of all the features. Our preliminary experiments showed no significant improvement after 75$\%$ hence we bound the upper limit to this value. We performed all the experiments on ARGO cluster (http://orc.gmu.edu) with dual Intel Xeon E5-2670 8 core CPUs and 64 GB memory. Source code implementation of the proposed algorithm discussed in this paper is made available at our website\footnote{https://cs.gmu.edu/$\sim$mlbio/featureselection} for repeatability and future use. 

\begin{table}[hbt]
\begin{centering} 
\caption{\textbf{Performance comparison of adaptive and global approach for feature selection based on Gini-Index with all features. LR + $l_1$-norm model is used for evaluation}} 
\label{table:AdaptiveFS} 
\begin{tabular}{|@{\hskip 0.02in} l@{\hskip 0.02in}| @{\hskip 0.02in}c@{\hskip 0.02in}| c| c|@{\hskip 0.02in} c@{\hskip 0.02in}|} 
\hline
\multirow{1}{*}{\bf{Dataset}} & \multirow{1}{*}{\bf{Metric}} & \multirow{1}{*}{\bf{Adaptive FS}} & \multirow{1}{*}{\bf{Global FS}} & \multirow{1}{*}{\bf{All Features}}\\
\hline
\multirow{2}{*}{{\bf{NG}}} & $\mu F_1$ & 76.16$\vartriangle$ & {\bf{76.39}}$\vartriangle$ & 74.94\\        
 & $MF_1$ & {\bf{76.10}}$\vartriangle$ & 76.07$\vartriangle$ & 74.56 \\       
\multirow{2}{*}{{\bf{CLEF}}} & $\mu F_1$ & {\bf{72.66}} & 72.27 & 72.17\\        
 & $MF_1$ & {\bf{36.73}}$\blacktriangle$ & 35.07$\vartriangle$ & 33.14\\       
\multirow{2}{*}{{\bf{IPC}}} & $\mu F_1$ &  {\bf{48.23}}$\blacktriangle$ & 46.35 & 46.14\\        
 & $MF_1$ & {\bf{41.54}}$\blacktriangle$ & 39.52 & 39.43\\   
\multirow{2}{*}{{\bf{DMOZ-SMALL}}} & $\mu F_1$ & {\bf{40.32}}$\vartriangle$ & 39.52 & 38.86\\        
 & $MF_1$ & {\bf{26.12}}$\blacktriangle$ & 25.07 & 24.77\\  
\multirow{2}{*}{{\bf{DMOZ-2010}}} & $\mu F_1$ & {\bf{35.94}} & 35.40 & 34.32 \\        
 & $MF_1$ & {\bf{23.01}} & 21.32 & 21.26\\  
\multirow{2}{*}{{\bf{DMOZ-2012}}} & $\mu F_1$ & {\bf{44.12}}& 43.94 & 43.92 \\        
 & $MF_1$ & {\bf{23.65}} & 22.18 & 22.13\\
\hline    
\end{tabular}
\par\end{centering}
    \begin{tablenotes}
      \small
      \item $\blacktriangle$ (and $\vartriangle$) indicates that improvements are statistically significant with 0.05 (and 0.1) significance level.
    \end{tablenotes}
\end{table}

\section{Results Discussion} 
\label{resDiscuss}
\subsection{{Case Study}}
To understand the quality of features selected at different internal nodes in the hierarchy we perform case study on NG dataset. We choose this dataset because we have full access to feature information. Figure \ref{FSfigure} demonstrates the results of top five features that is selected using best feature selection method $i.e.$, Gini-Index (refer to Figure \ref{figaccuracyL1} and \ref{figaccuracyL2}). We can see from the figure that selected features corresponds to the distinctive attributes which helps in better discrimination at particular node. For example, the features like \emph{Dod (Day of defeat or Department of defense)}, \emph{Car}, \emph{Bike} and \emph{Team} are important at node `{Rec}' to distinguish between the sub-class `{autos}', `{motorcycles}' and `{Sports}' whereas other features like \emph{Windows}, \emph{God} and \emph{Encryption} are irrelevant. This analysis illustrates the importance of feature selection for TD HC problem.

One important observation that we made in our study is that some of the features like \emph{Windows}, \emph{God} and \emph{Team} are useful for discrimination at multiple nodes in the hierarchy (associated with parent-child relationships). This observation conflicts with the assumption made in the work by Xiao et al. \cite{zhou2011hierarchical}, which attempts to optimize the objective function by necessitating the child node features to be different from the features selected at the parent node. 


\begin{table*}[t]
\centering 
\caption{\textbf{Comparison of memory requirements for LR + $l_1$-norm model}} 
\label{table:paramsize} 
\begin{tabular}{|l| r r r r r r r r|} 
\hline
\multirow{2}{*}{\bf{Dataset}} & \multicolumn{2}{c|}{{\bf{Adaptive FS}}} & \multicolumn{2}{c|}{{\bf{Global FS}}} & \multicolumn{2}{c|}{{\bf{All Features}}}\\
\cline{2-7}
 & {{\bf{\# parameters}}} & \multicolumn{1}{c|}{{\bf{size}}} & {{\bf{\# parameters}}} & \multicolumn{1}{c|}{{\bf{ size}}} & {{\bf{\# parameters}}} & \multicolumn{1}{c|}{{\bf{ size}}}\\
\hline
{\bf{NG}}& 982,805 & \multicolumn{1}{r|}{4.97 MB} & {\bf{908,820}} & \multicolumn{1}{r|}{{\bf{3.64 MB}}} & 1,652,076 & \multicolumn{1}{r|}{6.61 MB}\\
{\bf{CLEF}} & {\bf{4,715}} & \multicolumn{1}{r|}{{\bf{18.86 KB}}} & {{5,220}} & \multicolumn{1}{r|}{{{20.89 KB}}} & 6,960 & \multicolumn{1}{r|}{27.84 KB }\\
{\bf{IPC}} & {\bf{306,628,256}} & \multicolumn{1}{r|}{\bf{1.23 GB}} & {{331,200,000}} & \multicolumn{1}{r|}{{{1.32 GB}}} & 620,170,344 & \multicolumn{1}{r|}{2.48 GB}\\
{\bf{DMOZ-SMALL}} & {\bf{74,582,625}} & \multicolumn{1}{r|}{\bf{0.30 GB}} & {{85,270,801}} & \multicolumn{1}{r|}{{{0.34 GB}}} & 121,815,771 & \multicolumn{1}{r|}{0.49 GB}\\
{\bf{DMOZ-2010}} & {\bf{ 4,035,382,592}} & \multicolumn{1}{r|}{\bf{16.14 GB}} & {{4,271,272,967}} & \multicolumn{1}{r|}{{{17.08 GB}}} & 6,571,189,180 & \multicolumn{1}{r|}{26.28 GB}\\
{\bf{DMOZ-2012}} & {\bf{3,453,646,353}} & \multicolumn{1}{r|}{{\bf{13.81 GB}}} & {{3,649,820,382}} & \multicolumn{1}{r|}{{{14.60 GB}}} & 4,866,427,176 & \multicolumn{1}{r|}{19.47 GB}\\
\hline
\end{tabular}
\end{table*} 

\subsection{{Classification Performance Comparison}} 
\textbf{Global FS} - Figures \ref{figaccuracyL1} and \ref{figaccuracyL2} shows the $\mu F_1$ and M$F_1$ comparison of LR models with $l_1$-norm and $l_2$-norm regularization combined with various feature selection methods discussed in Section \ref{filterFeatSelect} respectively. We can see that all feature selection method (except Kruskal-Wallis) show competitive performance results in comparison to the full set of features for all the datasets. Overall, Gini-Index feature selection method has slightly better performance over other methods. MRMR methods have a tendency to remove some of the important features as redundant based on the minimization objective obtained from data-sparse leaf categories which may not be optimal and negatively influences the performance. The Kruskal-Wallis method shows poor performance because of the statistical properties that is obtained from data-sparse nodes \cite{2008danger}.  

On comparing the $l_1$-norm and $l_2$-norm regularized models of best feature selection method (Gini-Index) with all features, we can see that $l_1$-norm models have more performance improvement (especially for M$F_1$ scores) for all datasets whereas for $l_2$-norm models performance is almost similar without any significant loss. This is because $l_1$-norm assigns higher weight to the important predictor variables which results in more performance gain. 

Since, feature selection based on Gini-Index gives the best performance, in the rest of the experiments we have used the Gini-Index as the baseline for comparison purpose. Also, we consider $l_1$-norm model only due to space constraint.

\textbf{Adaptive FS} - Table \ref{table:AdaptiveFS} shows the LR + $l_1$-norm models performance comparison of adaptive and global approaches for feature selection with all features. We can see from the table that adaptive approach based feature selection gives the best performance for all the datasets (except $\mu F_1$ score of NG dataset which has very few categories). For evaluating the performance improvement of models we perform statistical significance test. Specifically, we perform sign-test for $\mu F_1$ \cite{yang1999re} and non-parametric wilcoxon rank test for $M F_1$. Results with 0.05 (0.1) significance level is denoted by $\blacktriangle$ ($\vartriangle$). Tests are between models obtained using feature selection methods and all set of features. We cannot perform test on DMOZ-2010 and DMOZ-2012 datasets because true predictions and class-wise performance score are not available from online web-portal.

Statistical evaluation shows that although global approach is slightly better in comparison to full set of features they are not statistically significant. On contrary, adaptive approach is much better with an improvement of $\sim$2\% in $\mu F_1$ and M$F_1$ scores which are statistically significant. 

\subsection{{Memory Requirements}} Table \ref{table:paramsize} shows the information about memory requirements for various models with full set of features and best set of features that are selected using global and adaptive feature selection. Upto 45$\%$ reduction in memory size is observed for all datasets to store the learned models. This is a huge margin in terms of memory requirements considering the models for large-scale datasets (such as DMOZ-2010 and DMOZ-2012) are difficult to fit in memory.

It should be noted that optimal set of features is different for global and adaptive methods for feature selection hence they have different memory requirements. Overall, adaptive FS is slightly better because it selects small set of features that are relevant for distinguishing data-sparse nodes present in CLEF, IPC and the DMOZ datasets. Also, we would like to point out that Table \ref{table:paramsize} represents the memory required to store the learned model parameters only. In practice, 2-4 times more memory is required for temporarily storing the gradient values of model paramaters that is obtained during the optimization process.  

\begin{table}
\begin{centering} 
\caption{\textbf{Feature selection preprocessing time (in minutes)($\downarrow$)}} 
\label{table:preprocessing} 
\begin{tabular}{|l| @{\hskip 0.02in}c@{\hskip 0.02in}|@{\hskip 0.02in} c @{\hskip 0.02in}|@{\hskip 0.02in}c@{\hskip 0.02in} |@{\hskip 0.02in} c@{\hskip 0.02in}|} 
\hline
\multirow{2}{*}{\bf{Dataset}} & \multicolumn{4}{c|}{\bf{Feature Selection Method}}\\
\cline{2-5}
& \bf{Gini-Index} & \bf{MRMR-D} & \bf{MRMR-Q} & \bf{Kruskal-Wallis}\\
\hline
\multirow{1}{*}{{\bf{NG}}} & {\bf{2.10}} & 5.33 & 5.35 & 5.42\\            
\multirow{1}{*}{{\bf{CLEF}}} & {\bf{0.02}} & 0.46 & 0.54 & 0.70\\            
\multirow{1}{*}{{\bf{IPC}}} & {\bf{15.24}} & 27.42 & 27.00 & 23.24\\        
\multirow{1}{*}{{\bf{DMOZ-SMALL}}} & {\bf{23.65}} & 45.24 & 45.42 & 34.65\\        
\multirow{1}{*}{{\bf{DMOZ-2010}}} & {\bf{614}} & 1524 & 1535 & 1314\\       
\multirow{1}{*}{{\bf{DMOZ-2012}}} & {\bf{818}}& 1824 & 1848 & 1268\\  
\hline    
\end{tabular}
\par\end{centering}
\end{table}

\subsection{{Runtime Comparison}} 
\textbf{Preprocessing Time} - Table \ref{table:preprocessing} shows
the preprocessing time needed to compute the 
feature importance using the 
different feature selection methods. The 
Gini-index method takes the least amount of time 
since it does not require the interactions 
between different features to rank the features. The 
MRMR methods are computationally 
expensive due to the large  number of 
pairwise comparisons between all the 
features to identify the redundancy information. On other hand, the 
Kruskal-Wallis method has overhead associated with 
determining ranking of each features with different classes.

\textbf{Model Training} - Table \ref{table:RuntimeL1Norm} shows the total training time needed for learning models. As expected, feature selection requires less training time due to the less number of features that needs to be considered during learning. For smaller datasets such as NG and CLEF improvement is not noticeable. However, for larger datasets with high-dimensionality such as IPC, DMOZ-2010 and DMOZ-2012 improvement is much higher (upto 3x order speed-up). For example, DMOZ-2010 dataset training time reduces from 6524 minutes to mere 2258 minutes.

\textbf{Prediction Time} - For the dataset with largest number of test instances, DMOZ-2012 it takes 37 minutes
to make predictions with feature selection as opposed to 48.24 minutes
with all features using the TD HC approach.

\begin{figure}
\centering
  \includegraphics[width=0.5\linewidth,height=3.0cm]{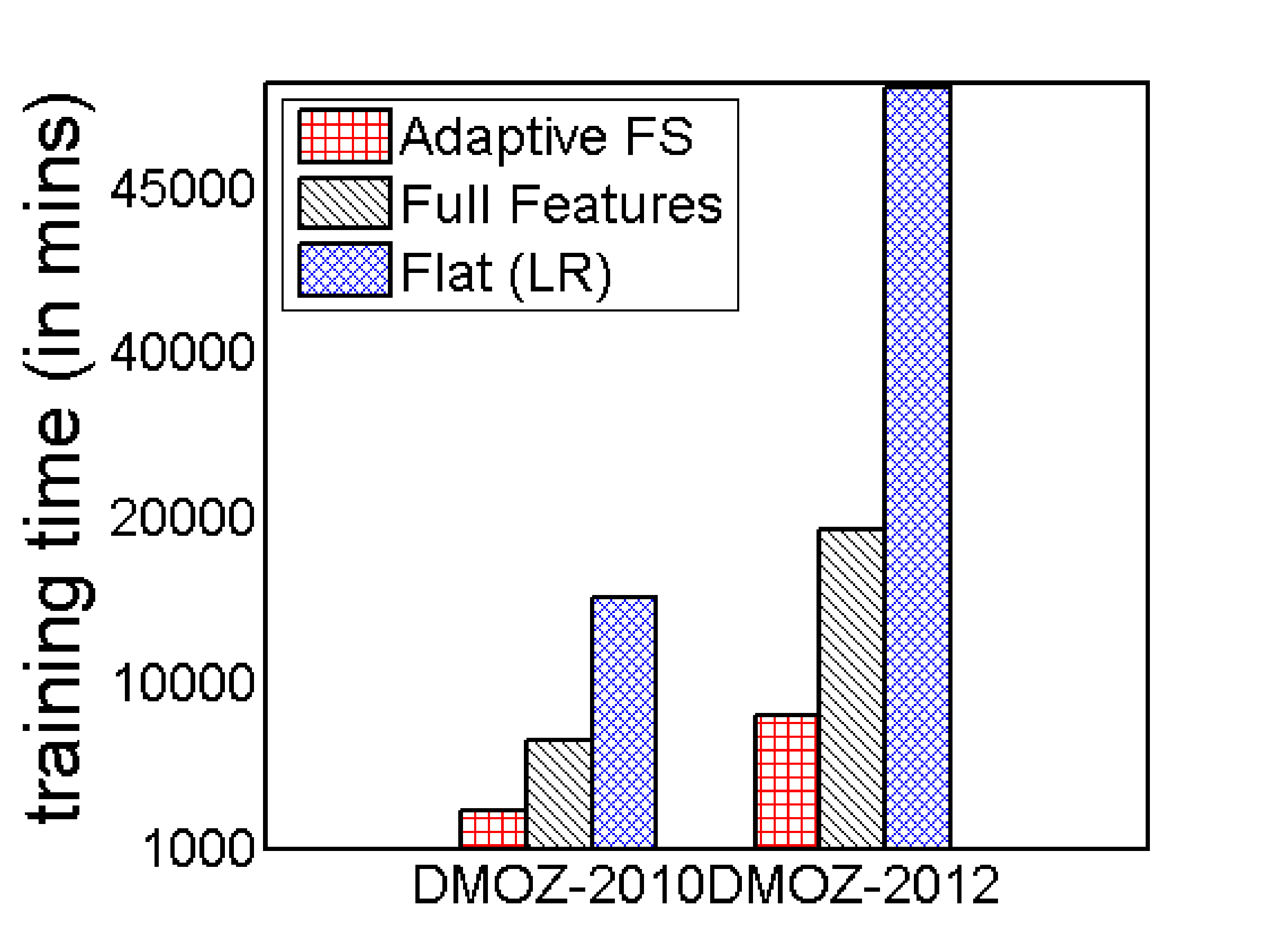}%
  \includegraphics[width=0.5\linewidth,height=3.0cm]{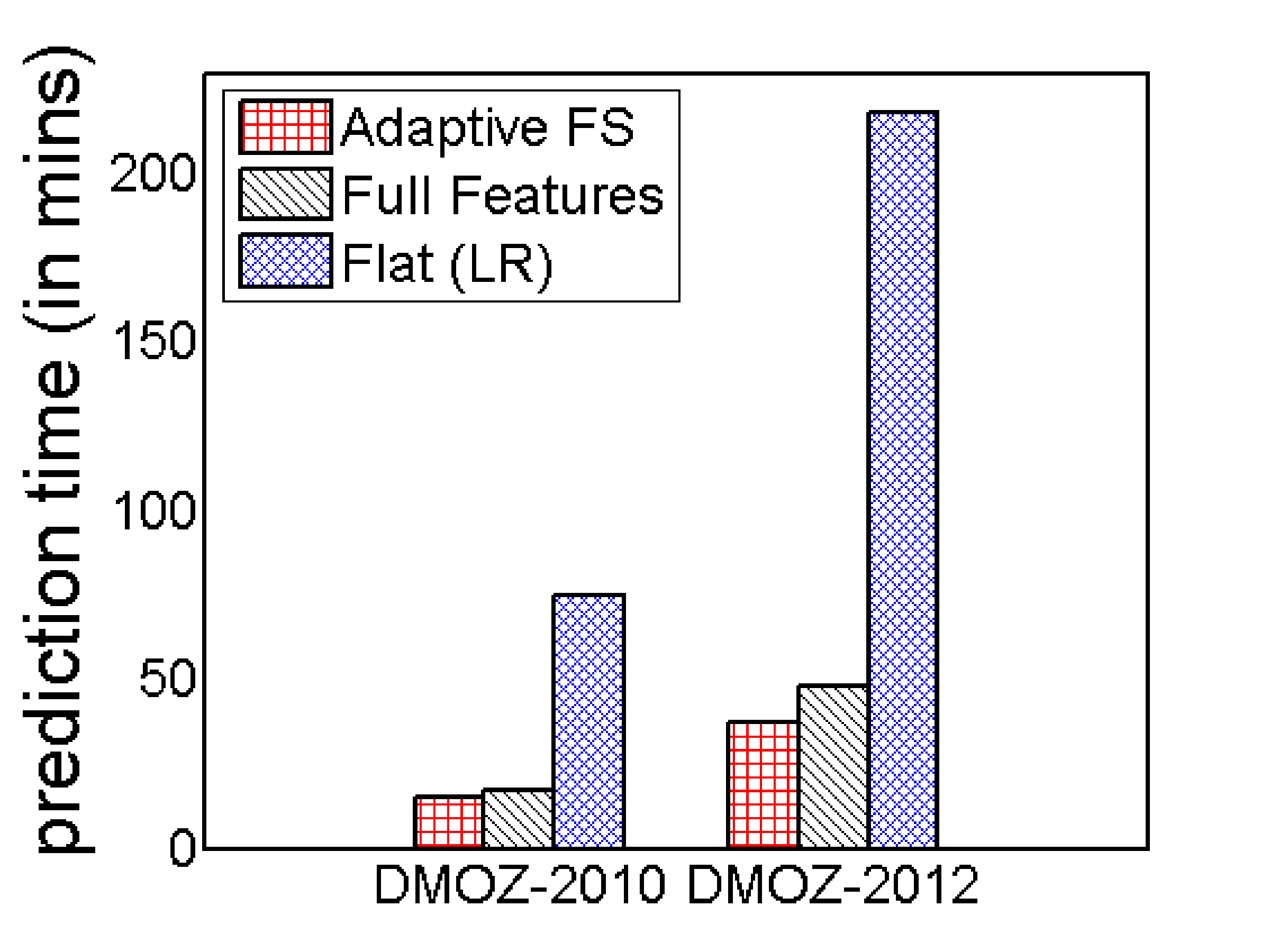}
  \caption{\textbf{Training and prediction runtime comparison of LR + $l_1$-norm model (in minutes).}}
\label{preComp}
\end{figure}

In Figure \ref{preComp} we  show the training and prediction time comparison of large datasets (DMOZ-2010 and DMOZ-2012) between flat LR and the TD HC approach with (and without) feature selection. The flat method is comparatively more expensive than the TD approach 
($\sim$6.5 times for training and $\sim$5 times for prediction). 

\begin{table}
\begin{centering} 
\caption{\textbf{Total training time (in minutes)($\downarrow$)}} 
\label{table:RuntimeL1Norm} 
\begin{tabular}{|l| c| c | c|} 
\hline
\multirow{2}{*}{\bf{Dataset}} & \multirow{2}{*}{\bf{Model}} & {\bf{Feature Selection}} & \multirow{2}{*}{\bf{All Features}}\\
&& {\bf{(Gini-Index)}} & \\
\hline
\multirow{2}{*}{{\bf{NG}}} & LR + $l_1$ & 0.75 & 0.94\\        
 & LR + $l_2$ & 0.44 & 0.69 \\       
\multirow{2}{*}{{\bf{CLEF}}} & LR + $l_1$ & 0.50 & 0.74\\        
 & LR + $l_2$ & 0.10 & 0.28\\       
\multirow{2}{*}{{\bf{IPC}}} & LR + $l_1$ &  24.38 & 74.10\\        
 & LR + $l_2$ & 20.92 & 68.58\\   
\multirow{2}{*}{{\bf{DMOZ-SMALL}}} & LR + $l_1$ &  3.25 & 4.60\\        
 & LR + $l_2$ & 2.46 & 3.17\\  
\multirow{2}{*}{{\bf{DMOZ-2010}}} & LR + $l_1$ & 2258 & 6524\\        
 & LR + $l_2$ & 2132 & 6418 \\  
\multirow{2}{*}{{\bf{DMOZ-2012}}} & LR + $l_1$ & 8024 & 19374\\        
 & LR + $l_2$ & 7908 & 19193\\  
\hline    
\end{tabular}
\par\end{centering}
\end{table}

\begin{table}
\begin{centering} 
\caption{\textbf{Performance comparison of LR + $l_1$-norm model with varying training size (\# instances) per class on NG dataset}}
\label{table:PerformanveVaryingTrainSize} 
\begin{tabular}{|@{}c@{}|@{}c@{}| c c|c c|} 
\hline
{\bf{Dataset}}& {\bf{Train size}} & \multicolumn{2}{c|}{\bf{Feature Selection}} & \multicolumn{2}{c|}{{\bf{All Features}}}\\
{\bf{Distribution}}& {\bf{(per class)}} & \multicolumn{2}{c|}{{\bf{(Gini-Index)}}} &{}&{}\\
\cline{3-6}
& & {\bf{$\mu F_1$}} & {\bf{M$F_1$}} & {\bf{$\mu F_1$}} & {\bf{M$F_1$}}\\
\hline
& \multirow{2}{*}{5} & {\bf{27.44}} $\blacktriangle$ 	& {\bf{26.45}} $\blacktriangle$ 	& 25.74	& 24.33\\
& & (0.4723)	& (0.4415)		& (0.5811)	& (0.6868)\\
& \multirow{2}{*}{10} & {\bf{37.69}} $\vartriangle$ 	& {\bf{37.51}} $\blacktriangle$  	& 36.59	& 35.86\\
{\bf{Low}}& & (0.2124)	& (0.2772)		& (0.5661)	& (0.3471)\\
{\bf{Distribution}}& \multirow{2}{*}{15} & {\bf{43.14}} $\vartriangle$ 	& {\bf{43.80}} $\vartriangle$ 	& 42.49	& 42.99\\
& & (0.3274)	& (0.3301)		& (0.1517)	& (0.7196)\\
& \multirow{2}{*}{25} & {\bf{52.12}} $\blacktriangle$ 	& {\bf{52.04}} $\blacktriangle$ 	& 50.33	& 50.56\\
& & (0.3962)	& (0.3011)		& (0.4486)	& (0.5766)\\
\hline
& \multirow{2}{*}{50} & {\bf{59.55}}	& 59.46	& 59.52	& {\bf{59.59}}\\
& & (0.4649)	& (0.1953)		& (0.3391)	& (0.1641)\\
& \multirow{2}{*}{100} & 66.53	& 66.42	& {\bf{66.69}}	& {\bf{66.60}}\\
{\bf{High}} & & (0.0346)	& (0.0566)		& (0.7321)	& (0.8412)\\
{\bf{Distribution}}& \multirow{2}{*}{200} & 70.60	& 70.53	& {\bf{70.83}}	& {\bf{70.70}}\\
& & (0.6068)	& (0.5164)		& (0.7123)	& (0.6330)\\
& \multirow{2}{*}{250}& 72.37	& 72.24	& {\bf{73.06}} $\vartriangle$ 	& {\bf{72.86}}  \\
& & (0.4285)	& (0.4293)		& (0.4732)	& (0.4898)\\
\hline
\end{tabular}
\par\end{centering}
    \begin{tablenotes}
      \small
      \item Table shows mean and (standard deviation) in bracket across five runs. $\blacktriangle$ (and $\vartriangle$) indicates that improvements are statistically significant with 0.05 (and 0.1) significance level.
    \end{tablenotes}
\end{table}

\subsection{{Additional Results}}
{\textbf{Effect of Varying Training Size}} - Table \ref{table:PerformanveVaryingTrainSize} shows the classification performance on NG dataset with varying training dataset distribution. We have tested the models by varying the training size (instances) per class ($t_c$) between 5 and 250. Each experiment is repeated five times by randomly choosing $t_c$ instances per class. Moreover, adaptive method with Gini-Index feature selection is used for experiments. For evaluating the performance improvement of models we perform statistical significance test (sign-test for $\mu F_1$ and wilcoxon rank test for $M F_1$). Results with 0.05 (0.1) significance level is denoted by $\blacktriangle$ ($\vartriangle$).

We can see from Table \ref{table:PerformanveVaryingTrainSize} that for 
low distribution datasets, 
the 
feature selection method performs well 
and shows 
improvements of upto 2$\%$ (statistically significant) over the baseline method. The  reason behind this 
improvement is that with low data distribution, feature selection methods 
prevents the models from overfitting by selectively 
choosing the important features that helps in 
discriminating between 
the models of various classes. 
For datasets with high distribution,
no significant performance gain is observed due to sufficient number of available 
training instances for learning models which prevents overfitting when using 
all the features. 

\begin{figure}
\centering
  \includegraphics[width=0.33\linewidth,height=2.75cm]{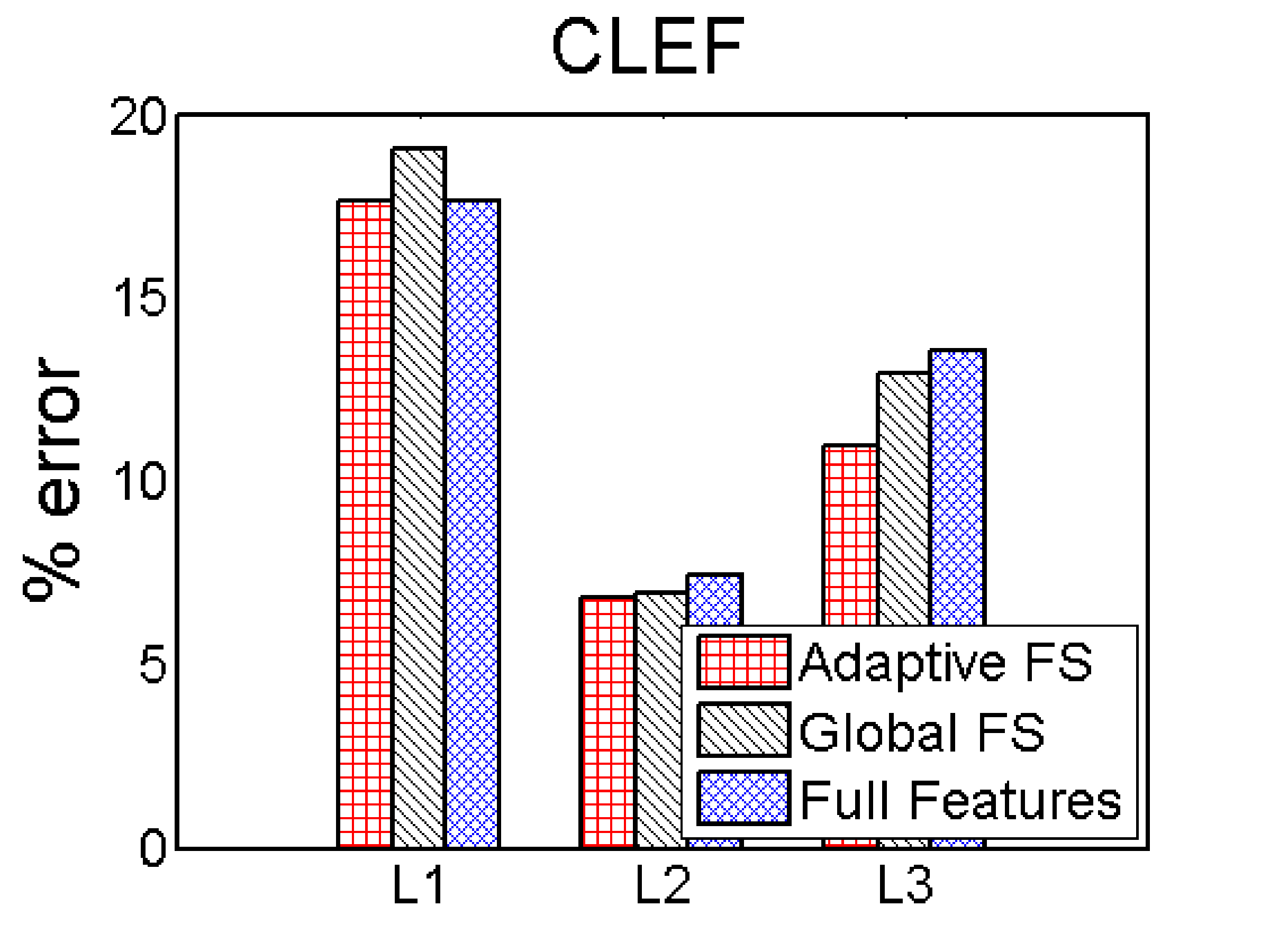}%
   \includegraphics[width=0.33\linewidth,height=2.75cm]{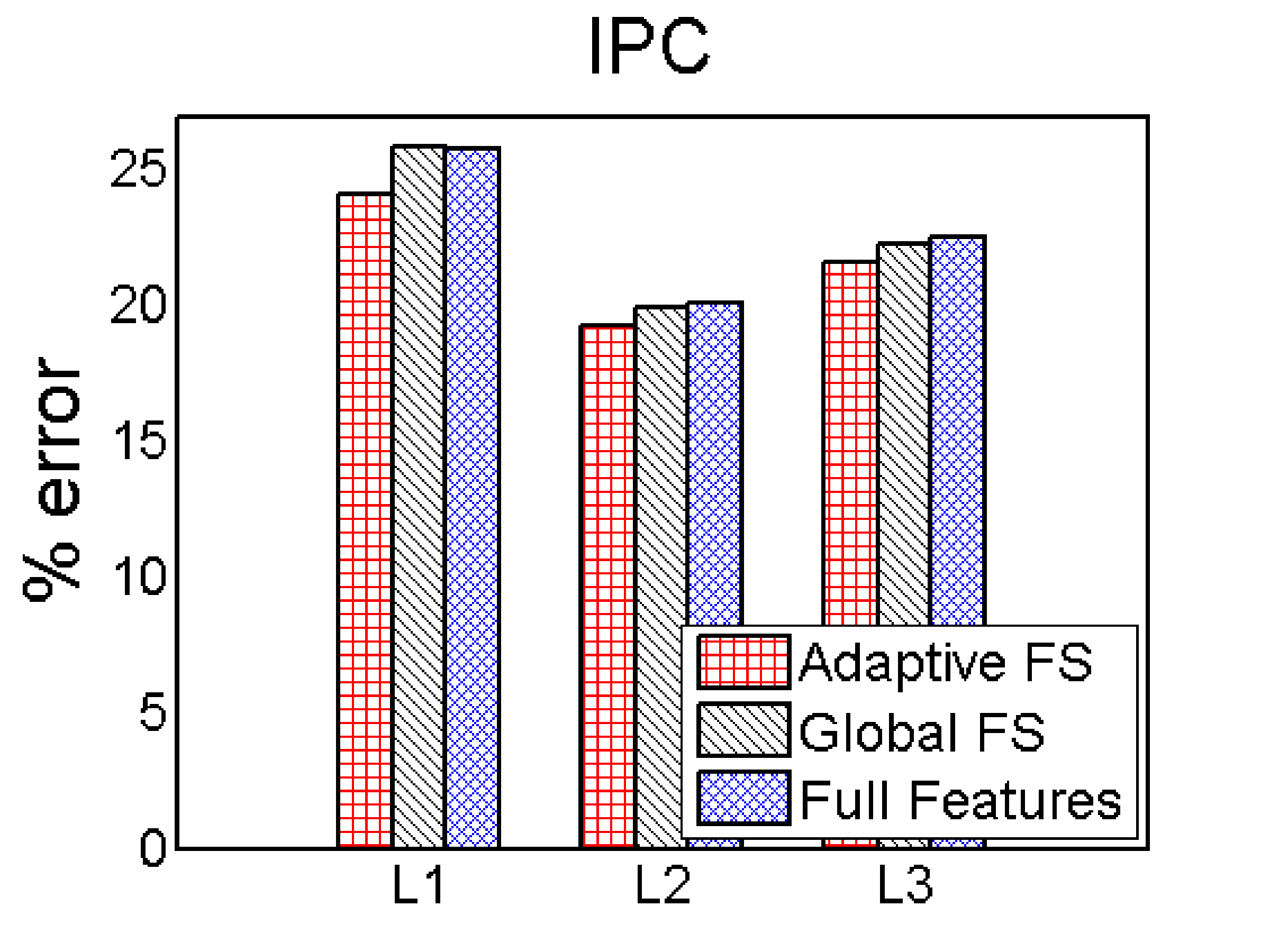}%
   \includegraphics[width=0.33\linewidth,height=2.75cm]{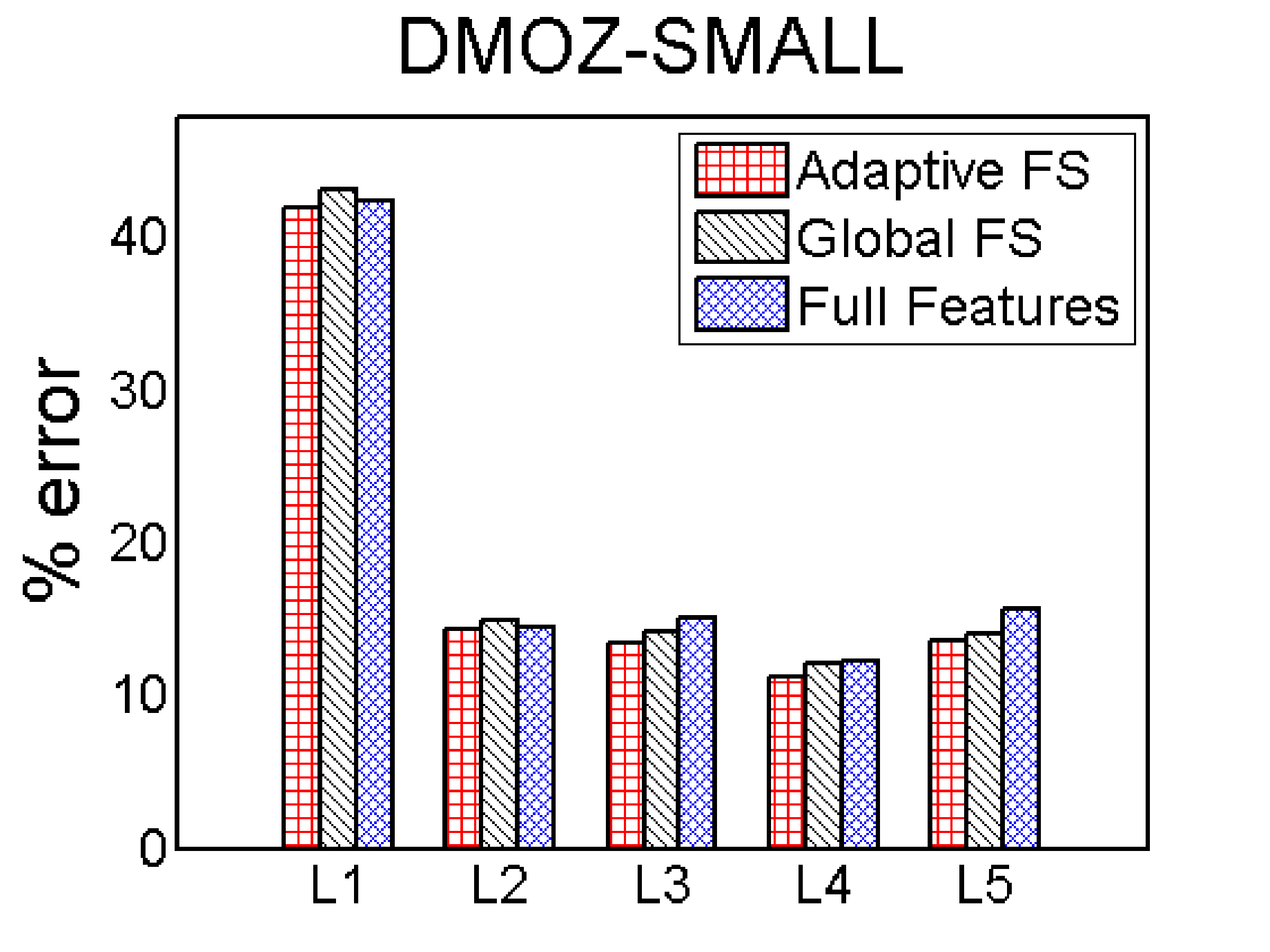}
  \caption{\textbf{Level-wise error analysis of LR + $l_1$-norm model for CLEF, IPC and DMOZ-SMALL datasets.}}
\label{LevelwiseCLEFfigure}
\end{figure}

{\textbf{Levelwise Analysis}} - Figure \ref{LevelwiseCLEFfigure} shows the level-wise error analysis for CLEF, IPC and DMOZ-SMALL datasets with or without feature selection. We can see that at topmost level more error is committed compared to the lower level. This is because at higher levels each of the children nodes that needs to be discriminated is the combination of multiple leaf categories which cannot be modeled accurately using the linear classifiers. Another observation is that adaptive feature selection gives best results at all levels for all datasets which demonstrates its ability to extract relevant number of features at each internal node (that belongs to different levels) in the hierarchy. 

\section{Conclusion and future work}
In this paper we compared various feature 
selection methods for solving large-scale HC problem. Experimental evaluation 
shows that with feature selection we are able to 
achieve significant improvement 
in terms of 
runtime performance (training and prediction) without 
affecting the accuracy of learned classification 
models. We also showed
that feature selection can be beneficial, especially 
for the larger datasets in terms of memory 
  requirements. This paper presents the first study of 
  various information theoretic feature selection 
  methods for large-scale HC.

In future, we plan to extend our work by learning more 
complex models at each of the decision nodes. Specifically, we 
plan to use multi-task learning methods where 
related tasks can be learned jointly to improve the 
performance on each task. Feature selection gives us 
the flexibility of
learning complex models due to 
reduced dimensionality of the features, which otherwise have longer runtime and larger memory requirements.

\section*{Acknowledgement}
NSF Grant \#1252318 and \#1447489 to Huzefa Rangwala and Summer Research Fellowship from the office of provost, George Mason University to Azad Naik.

\renewcommand{\bibfont}{\footnotesize}
\bibliographystyle{./IEEEtranBST/IEEEtran}
\bibliography{./IEEEtranBST/IEEEabrv,FeatureSelection_reference}

\end{document}